\PassOptionsToPackage{table}{xcolor} %
\documentclass[]{alaya}
\usepackage{makecell}
\usepackage{wrapfig}
\usepackage{tabularx}
\usepackage{textcomp}
\usepackage{stfloats}
\usepackage{url}
\usepackage{verbatim}
\usepackage{titlesec}
\usepackage{tocloft}
\usepackage{adjustbox}
\usepackage{multirow}
\usepackage{pifont}
\usepackage[sc]{mathpazo}
\usepackage{tikz}
\usepackage{comment}
\usepackage{amsmath,amssymb}
\usepackage{colortbl}
\usepackage[numbers,sort&compress]{natbib} %
\usepackage{booktabs} 
\usepackage{hyperref}
\usepackage{graphicx}
\usepackage{subcaption}
\usepackage{multirow}
\usepackage{subcaption}
\RequirePackage{xspace}
\makeatletter
\DeclareRobustCommand\onedot{\futurelet\@let@token\@onedot}
\def\@onedot{\ifx\@let@token.\else.\null\fi\xspace}
\usepackage[most]{tcolorbox}
\usepackage{array}
\usepackage{tabularx}
\usepackage{siunitx} 
\usepackage{makecell}
\usepackage[table]{xcolor}
\usepackage{caption}
\definecolor{headerpurple}{HTML}{d8d2fc}
\definecolor{rowgray}{gray}{0.95}

\makeatother

\definecolor{adptorange}{RGB}{248, 205, 172}
\definecolor{cmpblue}{RGB}{189, 215, 238}
\definecolor{cmpblue}{RGB}{189, 215, 238}

\definecolor{our_red}{RGB}{232,157,160}
\definecolor{our_blue}{RGB}{136,206,230}
\definecolor{our_orange}{RGB}{246,200,168}
\definecolor{our_green}{RGB}{178,211,164}

\definecolor{attn_code0}{RGB}{247,215,200}
\definecolor{attn_code1}{RGB}{238,169,139}
\definecolor{mlp_code0}{RGB}{204,201,221}
\definecolor{mlp_code1}{RGB}{102,95,153}
\definecolor{mygray}{HTML}{f0f0f0}

\definecolor{token_blue}{RGB}{84, 120, 140}

\newcommand{\sysname}{Sprint3D\xspace}
\newcommand{\Sysname}{Sprint3D\xspace}

\newif\ifdraftstatus
\draftstatusfalse   %

\definecolor{cmtkuno}{HTML}{E65100}     %
\definecolor{cmtyrh}{HTML}{7B1FA2}      %
\definecolor{cmtzhixiang}{HTML}{827717} %
\usepackage{CJKutf8} %

\usepackage{pifont}
\usepackage{bbding}
\usepackage{fontawesome}
\usepackage{xspace}

\usepackage{float}

\newlength\savewidth
\newcommand{\tablestyle}[2]{\setlength{\tabcolsep}{#1}\renewcommand{\arraystretch}{#2}\centering\footnotesize}

\newcolumntype{x}[1]{>{\centering\arraybackslash}p{#1pt}}
\newcolumntype{y}[1]{>{\raggedright\arraybackslash}p{#1pt}}
\newcolumntype{z}[1]{>{\raggedleft\arraybackslash}p{#1pt}}

\renewcommand{\paragraph}[1]{\vspace{1mm}\noindent\textbf{#1}}
\usepackage{colortbl}
\usepackage{wrapfig}

\renewcommand{\paragraph}[1]{\vspace{1.25mm}\noindent\textbf{#1}}

\usepackage{algorithm}
\usepackage{listings}

\definecolor{codeblue}{rgb}{0.25, 0.5, 0.5}
\definecolor{codekw}{rgb}{0.35, 0.35, 0.75}
\lstdefinestyle{Pytorch}{
    language = Python,
    backgroundcolor = \color{white},
    basicstyle = \fontsize{9pt}{8pt}\selectfont\ttfamily\bfseries,
    columns = fullflexible,
    aboveskip=1pt,
    belowskip=1pt,
    breaklines = true,
    captionpos = b,
    commentstyle = \color{codeblue},
    keywordstyle = \color{codekw},
}

\definecolor{green}{HTML}{009000}
\definecolor{red}{HTML}{ea4335}

\title{Sprint3D: Accelerating End-to-End 3D Asset Generation through Pipeline Optimization}
\author[1, 2]{Sho Kuno}
\author[1, 2]{Ruihan Yu}
\author[1]{Yonghao Yu}
\author[1]{Guixu Lin}
\author[1, 2]{Lian Fu}
\author[1, 2]{Muyao Niu}
\author[1, 3]{Yu-Ju Tsai}
\author[1]{Zheng-Hui Huang} %
\author[1, 5]{Yi-Chuan Huang} %
\author[1]{Siqi Yang} %
\author[1]{Fengbo Lan}
\author[1, 4]{Erwin Wu}
\author[1,\dagger]{Kaipeng Zhang}
\author[1, \dagger]{Zhixiang Wang} %

\affiliation[1]{Alaya Lab\\}
\affiliation[2]{The University of Tokyo}
\affiliation[3]{University of California, Merced}
\affiliation[4]{Institute of Science Tokyo}
\affiliation[5]{National Yang Ming Chiao Tung University}
\affiliation[\dagger]{Corresponding author} %

\abstract{
Image-to-3D generative models now produce asset-quality textured meshes, but at minutes per asset they remain slow to iterate with and costly to deploy.
Prior work has mostly accelerated the sampling of the networks, and where it also optimizes the mesh processing, it trains its own generators.
We instead start from a publicly released pipeline and ask how much of its end-to-end time can be saved.
Profiling the public Pixal3D pipeline shows that its latency is dominated by what happens after the networks: the four rectified flows take 14\% of the wall-clock time and mesh post-processing takes 85\%, most of it in CPU-based UV unwrapping.
We therefore first rebuild the mesh post-processing on the GPU and take it off the critical path, then cut the per-evaluation overhead of the flows and distill them from 76 to 8 network evaluations.
The resulting pipeline, \sysname, reduces the median time per asset from 133\,s to 4.05\,s on one H200 GPU at the same export settings (33$\times$).
At the same time, on 2,653 held-out Toys4K objects it improves every input-view normal metric of the Pixal3D protocol, with a slightly lower mean silhouette IoU, lower point-cloud alignment scores and less fine texture detail.
The rebuilt post-processing alone also speeds up TRELLIS.2, which uses the same export code, by 6.7$\times$ without a detectable change in the evaluated metrics.
}

\project{\url{https://alaya-lab.github.io/sprint3d/}}
\code{\url{https://github.com/AlayaLab/sprint3d}}
\date{\today}

\begin{document}
\maketitle

\begin{figure}[H]
    \centering
  \includegraphics[width=\linewidth]{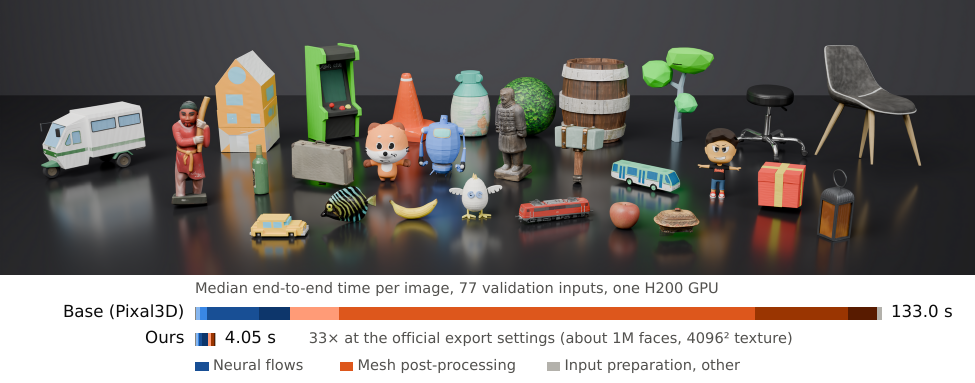}\vspace{5pt}
  \captionof{figure}{\textbf{Image-to-3D assets from our pipeline.} Twenty-eight assets, each generated from one image (validation objects; about 1M faces and a $4096^2$ texture per asset), rendered in Blender. Bottom: median end-to-end time, split by stage in proportion to the mean critical path.}
  \label{fig:teaser}
\end{figure}

\section{Introduction}
\label{sec:intro}

Image-to-3D generation has moved from proof of concept to production. Optimization-based methods introduced for text prompts distill a 2D diffusion prior into one object at a time and take minutes to hours per asset~\citep{rw_poole2023dreamfusion,rw_lin2023magic3d,rw_shi2023mvdream}. Novel-view and multi-view diffusion~\citep{rw_liu2023zero123,rw_shi2023zero123pp,rw_long2024wonder3d} and feed-forward reconstruction from one or a few views~\citep{rw_hong2024lrm,rw_xu2024instantmesh,rw_tang2024lgm,rw_tochilkin2024triposr,rw_boss2024sf3d} brought this down to minutes and then to seconds. Native 3D generation trains a diffusion or flow model directly on a scalable 3D latent representation and now produces textured meshes usable as game and product assets. Recent native generators (Section~\ref{sec:related}) raise the output resolution and surface detail, generate physically based materials~\citep{rw_hunyuan3d21} and follow the input image more closely~\citep{rw_li2026pixal3d}.

In production, however, such a generator is rarely run once: creators sample several candidates per input, compare them and re-run with adjusted inputs. The wall-clock time from an uploaded image to an exported asset therefore determines both how quickly they can iterate and how much GPU time each asset costs to serve. On an H200 GPU, the released pipelines that write a textured asset take from about one to more than two minutes per image in our measurements (Appendix Table~\ref{tab:app_baseline_e2e}).

Yet work on accelerating these generators has concentrated on the generative networks through few-step distillation, training-free caching and compression of the diffusion transformer, and it reports at most the latency of the accelerated network~\citep{rw_lai2025flashvdm,rw_zhou2025mdtdist,rw_sam3d2025,rw_yang2026fast3dcache,rw_lee2026vitality}. Where such work does profile a pipeline stage by stage, the profile stops at the decoded shape~\citep{rw_lai2025flashvdm,rw_feng2026fastsam3d}. Among the image-to-3D systems we surveyed, only AssetGen~\citep{rw_assetgen2026} breaks down the time from input image to exported asset by stage (Appendix~\ref{app:profile}).

Meanwhile, the three released pipelines we run that write a textured asset unwrap UVs with xatlas on the CPU~\citep{rw_xatlas}: Pixal3D~\citep{rw_li2026pixal3d}, TRELLIS.2~\citep{rw_xiang2025trellis2} and Hunyuan3D-2.1~\citep{rw_hunyuan3d21}. Prior work that also optimizes the steps after the networks does so together with its own generators~\citep{rw_boss2024sf3d,rw_assetgen2026}, which leaves open how much of a released pipeline's end-to-end time can be saved without training a new one. We therefore begin with a measurement and ask: \emph{where does the end-to-end time of a state-of-the-art image-to-3D pipeline go, and how much of it can be removed before the networks are touched?}

To answer it, we study Pixal3D~\citep{rw_li2026pixal3d}, whose public code and weights allow its end-to-end time to be both measured and changed. Its public implementation runs four rectified flows on the structured latents of TRELLIS.2~\citep{rw_xiang2025trellis2}, conditioned on the input pixels so that the output stays registered to the input view. At its official export (about 1M faces, $4096^2$ texture) it takes more than two minutes per image, and profiled stage by stage it spends most of this time in mesh post-processing and little in its four flows (Figure~\ref{fig:teaser}, bottom; Section~\ref{sec:method-overview}). TRELLIS.2 uses the same export code and shows the same split, and the two other released pipelines we timed also spend most of their time outside the sampling of their networks (Appendix Table~\ref{tab:app_baseline_e2e}). \textbf{The latency of such a pipeline is dominated by the work after its networks.}

The profile exposes three bottlenecks, each of which limits the latency once the previous one is removed, and our pipeline removes them in turn. The mesh post-processing comes first: it runs on the CPU and on the critical path, and UV unwrapping alone takes longer than all four flows together. We rebuild it on the GPU and take it off the critical path, with a single-pass GPU chart segmentation in place of the CPU unwrap. With the post-processing off the critical path, every network evaluation still has avoidable cost, which we remove by computing the conditioning once, reusing step-invariant keys and values, and using faster attention kernels. What remains is the number of evaluations: each flow takes 12 steps, and three of the four flows are guided. Reducing the steps without retraining degraded the quality. We therefore distill each flow separately and choose the objective and the step count per stage by ablation. Specifically, the coarse- and fine-shape flows are replaced by consistency-distilled students and the structure and texture flows by trajectory students, which reproduce in fewer steps the 12-step sampling trajectory of their teacher, the released flow. Because any of these optimizations can change the asset, we classify every change as output-identical, near-identical or output-changing and measure its effect end to end (Table~\ref{tab:optimizations}).

The result is \sysname, which writes the textured GLB at the official export in 4.05\,s at the median, compared with 133\,s for Base (Pixal3D), the released pipeline run with its own script (33$\times$; Figure~\ref{fig:teaser}). In the order in which the bottlenecks are removed, the speed-up splits into 7.3$\times$ for the rebuilt post-processing, 1.2$\times$ for the engineering of the unchanged flows and 3.8$\times$ for distillation (Table~\ref{tab:strong_baseline}). For quality, we follow the protocol of the Pixal3D paper: on 2,653 held-out Toys4K~\citep{ap_toys4k} test objects evaluated once, \sysname is significantly better than the released pipeline on all ten input-view normal columns, at a slightly lower mean silhouette IoU. Compared with three other open generators, \sysname and the released pipeline have the best values in every quality column of this protocol, whereas on the metrics outside it the comparison is mixed. The speed-up comes at the cost of lower point-cloud alignment scores and less fine texture detail than the 12-step teacher of the texture stage (Section~\ref{sec:comparison}). Finally, the rebuilt mesh post-processing is not specific to Pixal3D: without the half-resolution remesh it also transfers to TRELLIS.2, where the evaluated metrics do not change detectably (Table~\ref{tab:trellis2_meshpath}).

In summary, our contributions are:
\begin{itemize}
  \item An end-to-end profile of a state-of-the-art image-to-3D pipeline (Section~\ref{sec:method-overview}).
  \item A pipeline that removes the three bottlenecks in order, with a per-stage distillation cascade at non-uniform step counts. Its mesh post-processing transfers to TRELLIS.2.
  \item A decomposition of the end-to-end speed-up against two configurations that keep the 12-step flows at the official export, and an evaluation under Pixal3D's own protocol with the complete ablation, the measured trade-offs and the techniques we tried and rejected.
\end{itemize}

\section{Related Work}
\label{sec:related}

\paragraph{Native image-to-3D generation and its representations.}
Native generators differ in the latent they denoise. That latent is a set of latent vectors attached to the surface~\citep{rw_zhang2023vecset,rw_zhang2024clay,rw_li2025triposg,rw_hunyuan3d20,rw_li2025step1x3d}, a set of latent vectors anchored to a coarse voxel grid (LATTICE~\citep{rw_lai2025lattice}, the shape model of Hunyuan3D~2.5~\citep{rw_hunyuan3d25}), a triplane~\citep{rw_wu2024direct3d}, sparse voxels~\citep{rw_wu2025direct3ds2,rw_li2025sparc3d}, or structured latents on a sparse voxel grid~\citep{rw_xiang2025trellis}. %
TRELLIS introduced the structured latents, which are generated by a structure flow and then a latent flow over the active voxels and decoded into 3D Gaussians, radiance fields or meshes~\citep{rw_xiang2025trellis}. Hi3DGen conditions these structured latents on a normal map estimated from the image~\citep{rw_ye2025hi3dgen}, and TRELLIS.2 learns them from native 3D data with a compact sparse-voxel representation of geometry and material attributes, from which a mesh with material is decoded directly~\citep{rw_xiang2025trellis2}. The released Pixal3D pipeline, which we study, builds on TRELLIS.2 with pixel-aligned conditioning~\citep{rw_li2026pixal3d}. %
In contrast to native generation, feed-forward reconstruction returns an asset from one network pass in under a second: TripoSR outputs a vertex-colored mesh and SF3D a UV-unwrapped mesh with estimated material parameters~\citep{rw_tochilkin2024triposr,rw_boss2024sf3d}. %

\paragraph{Texture and material generation.}
Appearance is generated in two ways, and both end in a UV texture. Multi-view painting renders the mesh from several views, generates or inpaints the views with a 2D diffusion model and back-projects them into a UV texture~\citep{rw_richardson2023texture,rw_bensadoun2024texturegen,rw_cheng2024mvpaint}, as Hunyuan3D-2.1 does for its PBR material maps~\citep{rw_hunyuan3d21}. %
Native generators instead denoise an appearance latent on the same support as the geometry and bake the decoded attributes into a UV atlas~\citep{rw_xiang2025trellis2}. The released Pixal3D pipeline takes this second route, and its texture resolution is therefore determined by the bake rather than by the generator. %

\paragraph{From generated geometry to an exportable asset.}
Between the decoder and the file lie the steps of classical geometry processing. A surface is extracted from the decoded field~\citep{rw_ju2002dualcontouring}, simplified to a face budget~\citep{rw_garland1997qem}, cut into charts and parameterized~\citep{rw_levy2002lscm}, packed into an atlas~\citep{rw_xatlas}, and the attributes are baked into the atlas. %
GPU formulations exist for these steps: parallel simplification for low-poly modeling (PaMO)~\citep{rw_pamo2025}, real-time charting and packing for texture-space shading (FastAtlas), whose charts are the connected components of the visible triangles, parameterized by perspective projection~\citep{rw_fastatlas2025}, and interactive GPU atlas packing in rows (TABI)~\citep{rw_gu2026tabi}. Alternatively, texture representations defined on the surface itself, such as heat kernel textures, do without a UV atlas~\citep{rw_foti2026hktex}. %
Released pipelines integrate these steps after the generator and differ in how they unwrap. SF3D projects each face onto the cube face chosen by its normal instead of running xatlas, which can take up to 30\,s per asset~\citep{rw_boss2024sf3d}, and Hunyuan3D Studio predicts the UV seams with a learned autoregressive model~\citep{rw_hunyuan3dstudio2025}. AssetGen parallelizes xatlas instead and reports the time of each step, as well as the saving from each change in its ablation of the shape branch~\citep{rw_assetgen2026} (Appendix~\ref{app:profile}). %

\paragraph{Faster sampling: few-step distillation and training-free acceleration.}
The networks themselves are made faster by sampling in fewer steps or by making each step cheaper, and few-step students of a diffusion or flow model are trained in several ways. Progressive distillation halves the sampler's steps per round by regressing one student step onto two teacher steps~\citep{rw_salimans2022progressive}. Consistency models map any point of the probability-flow ODE to its endpoint~\citep{rw_song2023consistency}, consistency trajectory models extend this to arbitrary time pairs~\citep{rw_kim2024ctm}, and flow map matching casts all of them as learning the two-time flow map~\citep{rw_boffi2024flowmap}. Distribution-matching objectives match sample distributions instead~\citep{rw_yin2024dmd2}, and classifier-free guidance can be folded into the student~\citep{rw_meng2023guided}. More recently, CrossDistill applies a trajectory-preserving objective at high noise and distribution matching on the student's own samples at low noise~\citep{rw_liu2026crossdistill}. PDD is a trajectory-based method like the trajectory students used here: it predicts several denoising steps per network evaluation and is trained on the student's own outputs, as CrossDistill is~\citep{rw_shaul2026pdd}. %
In 3D, FlashVDM distills the shape flow of Hunyuan3D-2 from 50 to 5 steps and speeds up its decoder~\citep{rw_lai2025flashvdm}, MDT-dist distills both TRELLIS flows to one or two steps each~\citep{rw_zhou2025mdtdist}, and SAM 3D cuts its geometry flow from 25 to 4 evaluations~\citep{rw_sam3d2025}. AssetGen applies progressive distillation stage by stage, with guidance folded into the shape students and also into the texture student in its faster configuration AssetGen Flash~\citep{rw_assetgen2026}. Seed3D~2.0 also folds guidance into all its geometry and texture transformers and then halves their sampling steps round by round~\citep{rw_gu2026seed3d2}. Outside 3D, TurboT2VA pairs consistency distillation with a quantized, fused inference stack for text-to-video-audio generation~\citep{rw_yang2026turbot2va}. %
The cost of sampling also falls without training, through higher-order solvers~\citep{rw_lu2022dpmsolver}, feature caching across steps~\citep{rw_ma2023deepcache,rw_yang2026fast3dcache,rw_feng2026fastsam3d} and token reduction~\citep{me_bolya2023tome,me_bolya2023tomesd,rw_feng2026fastsam3d}, while pruning, quantization and fine-tuning compress the diffusion transformer itself~\citep{rw_lee2026vitality}. %
Finally, FlashAttention-3~\citep{rw_shah2024flashattention3}, CUDA graphs and stream overlap are the standard tools of diffusion serving at the kernel level. %

\paragraph{Relation to this report.}
Against this background, we count the mesh post-processing as part of the latency, as AssetGen does (Section~\ref{sec:method-infra-mesh}), and we distill each stage separately, as MDT-dist and AssetGen do (Section~\ref{sec:method-distill}). Unlike AssetGen, which trains its own generator, we keep the released networks frozen and train low-rank adapters only (Section~\ref{sec:method-training}). %

\section{Method}
\label{sec:method}
\begin{figure*}[!t]
\centering
\includegraphics[width=\linewidth]{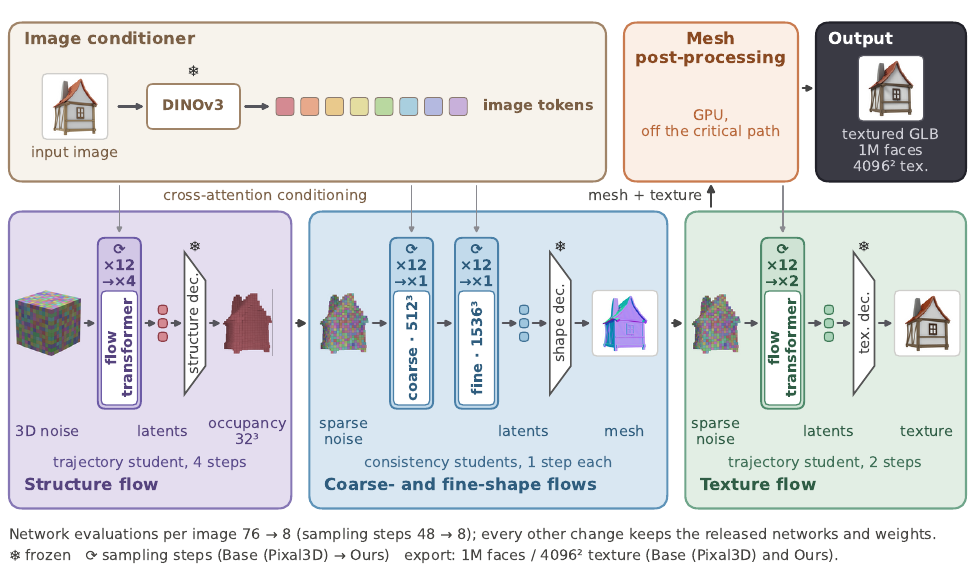} %
\caption{\textbf{Our pipeline.} The four flows of Pixal3D in three panels, with their sampling steps (Base (Pixal3D) $\to$ Ours) and the changes we apply.}
\label{fig:pipeline}
\end{figure*}

\subsection{Overview}
\label{sec:method-overview}

\paragraph{The Pixal3D pipeline.}
We study the released implementation of Pixal3D~\citep{rw_li2026pixal3d}, which turns one image into a textured mesh with a cascade of four rectified-flow stages~\citep{me_liu2023rectified,me_lipman2023flow} on the structured latents of TRELLIS.2~\citep{rw_xiang2025trellis2} (Figure~\ref{fig:pipeline}). In contrast, the model of the Pixal3D paper is built on Direct3D-S2~\citep{rw_wu2025direct3ds2} and generates the geometry only. In the released implementation, the input is cropped to the object and its field of view is estimated by MoGe-2~\citep{me_wang2025moge2}. A frozen DINOv3 encoder~\citep{me_simeoni2025dinov3} conditions every stage with features sampled where each token's voxel projects into the input image. Of the four stages, the \emph{structure flow} denoises a dense $16^3$ latent that a frozen decoder turns into a $64^3$ occupancy grid, max-pooled to a $32^3$ support. Its active voxels become the token set of the \emph{coarse-shape flow}, and after upsampling they become that of the \emph{fine-shape flow}. Both shape flows operate on sparse voxels, and the \emph{texture flow} produces the appearance latent on the same sparse tokens. Frozen VAE decoders turn the shape and texture latents into a dual grid with per-voxel attributes. A mesh path then turns that into a GLB: mesh extraction, remeshing, simplification, UV unwrapping, texture baking and file export. The released code builds this path on the CuMesh library~\citep{rw_xiang2025cumesh} and the nvdiffrast rasterizer~\citep{rw_laine2020nvdiffrast}.

\paragraph{Sampling.}
Each of the four flows is sampled with 12 Euler steps. The structure flow and the two shape flows use classifier-free guidance~\citep{me_ho2022cfg} at high noise levels ($t\geq0.6$), where a step costs two network evaluations, and the texture flow is unguided. In total, the released sampler performs 76 network function evaluations per image ($22+21+21+12$ for the four flows). We call the released pipeline run with its public inference script at its defaults \emph{Base (Pixal3D)} (Section~\ref{sec:setup}), and the released flow of one stage sampled in this way the \emph{12-step teacher} of that stage.

\paragraph{Where the time goes.}
To profile Base (Pixal3D), we measure the \emph{end-to-end (E2E) time} of a request from the input image to the written GLB (Section~\ref{sec:setup}). A second in which stages run concurrently is attributed to the stage the request is waiting for, the critical path. Measured this way, Base (Pixal3D) spends most of its time after the networks: in the dedicated timing run, mesh post-processing takes 85\% of the mean E2E time and the four flows with their conditioning 14\% (Appendix Table~\ref{tab:app_profile}). Most of that post-processing is spent in a single step, UV unwrapping (a GPU chart clustering followed by the CPU unwrapper xatlas). This step takes 94\,s on average, several times as long as the four flows together. The rest of the post-processing consists of the $4096^2$ texture bake, mesh extraction and remeshing with the BVH build, and the WebP export. Among the flows, the fine-shape flow takes the largest share, followed by the texture flow.

\paragraph{Optimizations and how they are measured.}
Guided by this profile, we remove three bottlenecks in Sections~\ref{sec:method-infra}--\ref{sec:method-model}, in the order in which they limit the E2E time: the CPU mesh post-processing on the critical path, the per-evaluation cost of the flows and the number of evaluations. Table~\ref{tab:optimizations} lists the adopted optimizations in these three groups, each with its output class. An optimization is \emph{output-identical} (=) when it is designed to leave the output unchanged and \emph{near-identical} ($\approx$) when its output differs by bf16 accumulation order, rounding or non-deterministic kernels, whereas distilled students, approximate geometry or a different UV chart layout make it \emph{output-changing} ($\blacklozenge$). We measure the effects in the \emph{cumulative measurement} on the 40-object subset of the validation set (80 inputs; Section~\ref{sec:setup}, Figure~\ref{fig:waterfall}). It starts from the re-run of Base (Pixal3D) inside our measurement code and adds the optimizations one at a time in the order of the three groups. In this order, the students are added last, so the released 12-step flows remain until then.

\subsection{Rebuilding the Mesh Post-Processing}
\label{sec:method-infra}
\label{sec:method-infra-mesh}
\label{sec:method-infra-pipeline}
The first bottleneck is the CPU mesh post-processing on the critical path. We therefore replace the released GLB export by the tiled mesh path, add seven optimizations to it and call the result the \emph{rebuilt mesh path}. Throughout, the networks and the generated latents are untouched, and the seconds quoted in this section are E2E effects from Table~\ref{tab:optimizations} unless stated otherwise.

\begin{figure}[!t]
\centering
\includegraphics[width=\linewidth]{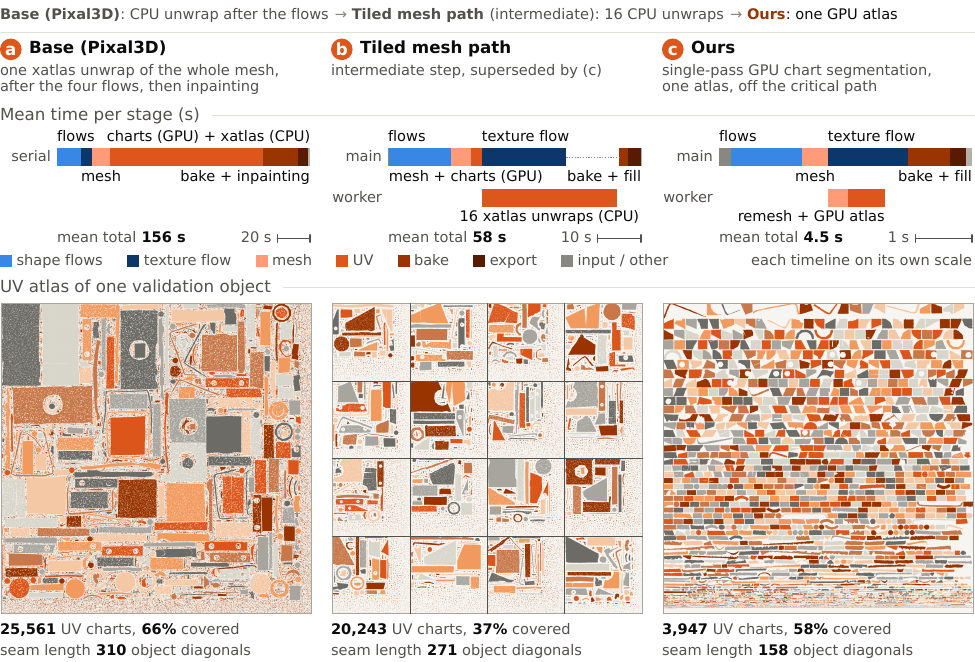}
\caption{\textbf{Three ways to build the UV atlas.} (a)~Base (Pixal3D); (b)~the tiled mesh path, an intermediate step; (c)~\sysname, with the GPU chart segmentation. Top: mean time per stage, two lanes for concurrent work (Appendix Table~\ref{tab:app_profile}). Bottom: the atlas of one validation object, charts in random tones, with chart count, texel coverage and seam length (object diagonals).}
\label{fig:atlas_paths}
\end{figure}

\paragraph{Tiled mesh path ($\blacklozenge$).}
The \emph{tiled mesh path} is the first form of our export and makes two changes to the UV unwrapping, the single step that takes most of the post-processing of the released path. First, we split the export into a geometry half and a bake half, so that the shape is decoded first and unwrapped on a thread while the texture flow samples. Second, we parallelize the unwrap itself: large charts are cut, and the charts are assigned to 16 groups, each unwrapped by its own instance of the CPU unwrapper xatlas~\citep{rw_xatlas} and packed into one tile of a $4\times4$ atlas. With the charts cut and tiled, the UV layout changes. However, the tiled mesh path is an intermediate configuration: \sysname replaces its per-tile CPU unwrapping by the single-pass GPU chart segmentation described below, and we keep it here because it is the first step of the cumulative measurement.

\begin{table}[!t]
\centering
\caption{\textbf{The adopted optimizations}, in the order of the cumulative measurement; a later row can replace an earlier one. Class: = output-identical, $\approx$ near-identical, $\blacklozenge$ output-changing. Effect: paired mean change of the E2E time when the row is added [95\% CI over the inputs]. Verification of the = and $\approx$ rows: Appendix~\ref{app:profile}.} %
\label{tab:optimizations}
\tablestyle{3pt}{1.1}
\begin{tabular}{@{}llcl l@{}}
\toprule
Optimization & Stage & Class & Effect (s) & Trade-off \\
\midrule
\multicolumn{5}{@{}l}{\textit{(1) Mesh post-processing (Section~\ref{sec:method-infra})}} \\
Tiled mesh path (16-tile atlas) & mesh post-proc. & $\blacklozenge$ & $-$116.1 [$-$141.3, $-$94.4] & 16-tile UV layout; superseded \\
Mesh pre-processing on a worker thread & mesh $\parallel$ texture & = & $-$0.20 [$-$1.46, +1.08] & scheduling only \\
Fixed bake-kernel launch configuration & bake & = & $-$1.91 [$-$3.19, $-$0.66] & identical in a unit test \\
Faster GPU simplification and charting & simplify, charts & $\approx$ & $-$0.71 [$-$2.10, +0.69] & non-deterministic, in spread \\
GPU BVH build & BVH & $\approx$ & $-$1.66 [$-$2.79, $-$0.52] & ties change nearest face \\
Adaptive half-resolution remesh & remesh & $\blacklozenge$ & +2.83 [+1.27, +4.44] & intermediate geometry changes \\
Lean GLB writer, parallel PNG & export & = & $-$2.68 [$-$4.07, $-$1.31] & only file bytes differ \\
Single-pass GPU chart segmentation & UV atlas & $\blacklozenge$ & $-$30.5 [$-$33.7, $-$27.4] & chart layout changes \\
\midrule
\multicolumn{5}{@{}l}{\textit{(2) Flow engineering (Section~\ref{sec:method-infra-flows})}} \\
FlashAttention-3, texture flow & texture & $\approx$ & $-$1.86 [$-$2.37, $-$1.45] & bf16 accumulation order \\
Texture K/V reuse across steps & texture & = & $-$0.16 [$-$0.19, $-$0.13] & identical texture latents \\
Shared image conditioning & conditioning & = & $-$0.41 [$-$0.44, $-$0.38] & identical tensors in a test \\
FlashAttention-3, fine-shape flow only & fine shape & $\approx$ & $-$2.84 [$-$3.43, $-$2.28] & active voxels unchanged \\
Side-stream texture sampling, early PNG & scheduling & = & $-$0.075 [$-$0.122, $-$0.035] & scheduling only \\
Texture projection reuse across steps & texture & = & $-$0.26 [$-$0.34, $-$0.19] & identical texture latents \\
\midrule
\multicolumn{5}{@{}l}{\textit{(3) Distilled students (Section~\ref{sec:method-model})}} \\
Structure student: trajectory, 4 steps & structure & $\blacklozenge$ & $-$0.92 [$-$1.48, $-$0.50] & IoU to ground truth unchanged \\
Shape students: consistency, 1 step each & coarse, fine shape & $\blacklozenge$ & $-$9.06 [$-$10.18, $-$8.01] & no loss in the 3D metrics \\
LoRA folded, fine-shape student & fine shape & $\approx$ & $-$0.17 [$-$0.20, $-$0.13] & bf16 rounding \\
Texture student: trajectory, 2 steps & texture & $\blacklozenge$ & $-$3.58 [$-$4.09, $-$3.10] & beats the same-step teacher \\
\bottomrule
\end{tabular}

\end{table}

\paragraph{Remeshing and simplification ($\blacklozenge$, $\approx$).}
Before the unwrap, the released path remeshes the decoded dual grid with narrow-band dual contouring~\citep{rw_ju2002dualcontouring} at full resolution and then simplifies to the face budget by parallel edge collapse with quadric error costs~\citep{rw_garland1997qem,rw_pamo2025}, both in CuMesh. To shorten this step, our \emph{adaptive half-resolution remesh} halves the remesh resolution and thereby changes the intermediate geometry before simplification. The exception is a fine-shape token set that is dense for the grid resolution, as for thin, detailed structures. Because the remesh extracts a thin double-walled shell whose wall is two grid cells thick, the halved grid also doubles the thickness of the never-rendered inner wall. This does not change any rendered metric but lowers the point-cloud alignment scores (Appendix~\ref{app:perception_gap}). The second change here is a faster GPU simplification and chart-clustering kernel, which is non-deterministic but stays within the run-to-run spread of the kernel it replaces.

\begin{figure}[!tb]
\centering
\includegraphics[width=\linewidth]{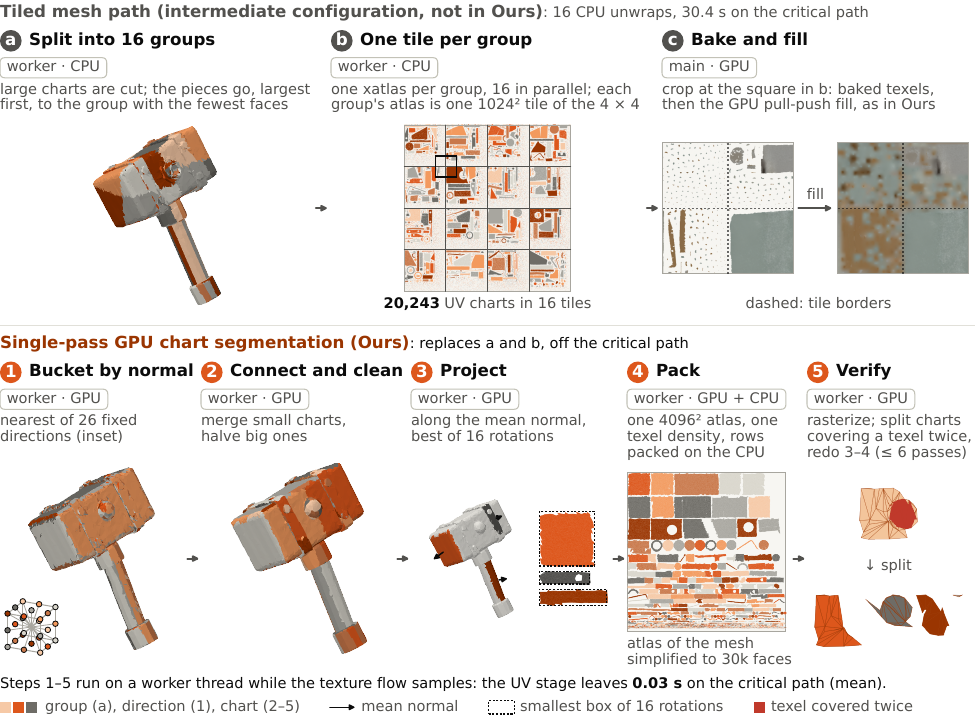}
\caption{\textbf{Two ways to build the UV atlas, step by step.} One validation object, shown simplified to 30k faces. Top: the tiled mesh path (not in \sysname): charts assigned to 16 groups (a), each unwrapped by xatlas into one tile of a $4\times4$ atlas (b); GPU pull-push fill (c). Bottom: the single-pass GPU chart segmentation (1--5), replacing (a) and (b).}
\label{fig:chartseg}
\end{figure}

\paragraph{BVH and UV atlas ($\approx$, $\blacklozenge$).}
The released path also builds the bounding-volume hierarchy used for attribute look-ups with cuBVH~\citep{rw_tang2022cubvh} on the CPU. Instead, we build it on the GPU as a linear BVH~\citep{rw_karras2012lbvh}, which leaves the distances unchanged and changes the nearest face only on exact ties ($\approx$). Since the tiled mesh path leaves the CPU unwrapping on the critical path, we replace its 16-tile CPU unwrapping altogether by a \emph{single-pass GPU chart segmentation} (Figure~\ref{fig:atlas_paths}). It finds the charts in one pass without iterative chart clustering, parameterizes them by a per-chart projection on the GPU and packs them into a single atlas, leaving the geometry unchanged but not the chart layout ($\blacklozenge$) (Appendix Figure~\ref{fig:chartseg}). Specifically, each face takes the nearest of a fixed set of directions, and the resulting charts are merged or cut according to their size (steps and parameters: Appendix~\ref{app:profile}). These parts follow earlier work: the fixed projection directions follow the box projection of SF3D~\citep{rw_boss2024sf3d}, the charts found as connected components follow FastAtlas~\citep{rw_fastatlas2025}, and the row packing follows the GPU packers of FastAtlas and TABI~\citep{rw_gu2026tabi}.
The chart segmentation removes 30.5\,s, the largest saving after the tiled mesh path itself: the remaining CPU unwrapping leaves the critical path, and the 12-step texture flow becomes the longer branch.

\paragraph{Baking and export (=).}
In the subsequent bake and export, two changes leave only the file bytes different: the Triton kernel that interpolates attributes into the texture uses a fixed launch configuration instead of autotuning at request time, and a lean GLB writer with parallel PNG encoding replaces the generic exporter. The decoded pixels and the geometry buffer are identical, and each change saves a few seconds (Table~\ref{tab:optimizations}).

\paragraph{Worker thread (=).}
Finally, the overlap with the texture flow is not limited to the UV atlas. A worker thread starts as soon as the shape latent is decoded and runs the whole geometry pre-processing (hole filling, BVH build, remeshing, simplification and the GPU atlas) on the GPU while the texture flow samples. This scheduling leaves the arithmetic unchanged.

\paragraph{Measured effect.}
What each of these optimizations saves depends on what is on the critical path when it is added. With the released 12-step flows, the tiled mesh path alone reduces the median E2E time of the re-run of Base (Pixal3D) from 138.4 to 55.3\,s. However, its 16-tile CPU unwrapping hides the 12-step texture flow on most inputs and limits the E2E time. At its position in the order, the adaptive half-resolution remesh lengthens the E2E time ($+2.8$\,s, Table~\ref{tab:optimizations}) since the CPU unwrapping is still active there and takes longer on the critical path than the mesh stage saves. The same remesh shortens the E2E time ($-0.83$\,s in a control run, Appendix~\ref{app:profile}) in the rebuilt mesh path, where the GPU chart segmentation has replaced the CPU unwrapping. %
 The worker thread contributes little: switching it off in \sysname lengthens the E2E time by only 0.03\,s, because the concurrent GPU work slows the texture flow about as much as it hides (Appendix~\ref{app:profile}).

\subsection{Engineering the Flows}
\label{sec:method-infra-flows}
Once the post-processing is off the critical path, the released 12-step flows limit the E2E time (Table~\ref{tab:strong_baseline}). Before shortening them we remove the second bottleneck, the cost that does not depend on the number of steps. All changes below leave the networks unchanged.

\paragraph{Reuse (=) and attention kernels ($\approx$).}
Part of this cost is repeated work. Since the released pipeline re-encodes the input image for every stage, we run the DINOv3 encoder once for each of the two input resolutions that the stages use, project its features once to the sparse token layout and share the upsampling encoder across the four flows. The conditioning tensors are identical. A second kind of repeated work occurs inside the texture flow: the cross-attention keys and values of the image tokens, and the context projections that produce them, do not depend on the sampling step and are thus computed at the first step and reused. The reuse of the projections can be switched on only together with the structure student and is therefore counted toward the distillation factor.
The near-identical changes are the attention kernels: the texture flow and the fine-shape flow use FlashAttention-3~\citep{rw_shah2024flashattention3} in bf16, an exact-attention kernel whose results differ from the reference implementation only by floating-point rounding, not by the algorithm. The structure and coarse-shape flows keep their attention kernels: the structure stage selects its active voxels by thresholding, so bf16-level perturbations there flip voxels and produce a different sample. Among the shape stages, restricting the changes to the fine-shape flow likewise keeps the active voxels identical and retains most of the gain.
FlashAttention-3 gives most of the gain of this section (Table~\ref{tab:optimizations}).

\paragraph{Scheduling (=).}
Three scheduling changes complete the flow engineering: texture sampling runs on a side CUDA stream synchronized by events, PNG encoding of the finished texture images starts while baking is still running, and the cache release is deferred to the end of the request. Further re-scheduling and prefetching did not reduce the E2E time (Appendix~\ref{app:ledger}).

\subsection{Per-Stage Distillation}
\label{sec:method-model}
\label{sec:method-distill}
What remains is the third bottleneck, the number of network evaluations, which we reduce by distilling each of the four flows into a student that takes fewer steps.

\paragraph{Setting.}
Each stage is a rectified flow: for a clean latent $x_0$ and Gaussian noise $\epsilon$ the network
$v_\theta(x_t,t,c)$ is trained to predict $\epsilon-x_0$ at $x_t=(1-t)\,x_0+t\,\epsilon$, $t\in[0,1]$, given the image
conditioning $c$. Sampling integrates $\mathrm{d}x_t=v\,\mathrm{d}t$ from $t=1$ to $0$, and the 12-step teacher takes
its Euler steps on a time grid $1=t_0>t_1>\dots>t_{12}=0$ that is shifted toward high noise.
Every student is a rank-256 LoRA adapter~\citep{me_hu2022lora} on the frozen
network of its teacher, initialized at zero so that the student equals the teacher at the start of training. An exponential moving average
of the adapter is deployed (batch sizes and decay: Appendix~\ref{app:distill}). We use two objectives.

\paragraph{Consistency distillation (coarse- and fine-shape flows).}
The first objective is consistency distillation~\citep{rw_song2023consistency} from a guided teacher as in latent consistency models~\citep{rw_luo2023lcm}, and we use it for the two shape flows in an ODE-anchored variant (the noisy states below are drawn around the guided teacher's endpoint, not around data). Let
$f_\theta(x_t,t,c)=x_t-t\,v_\theta(x_t,t,c)$ be the one-jump estimate of the clean latent. For each training view we
run the guided 12-step teacher once and keep its endpoint $\hat{x}_0$. Noisy states are then drawn on the path
$x_\tau=(1-\tau)\,\hat{x}_0+\tau\,\epsilon$ at the points $\tau_{n+1}$ of a 50-point discretization of
$[0,1]$. The student is trained to be consistent with the teacher's flow,
\begin{equation}
\mathcal{L}_{\mathrm{CD}}=\mathbb{E}\Big[\big\|f_\theta(x_{\tau_{n+1}},\tau_{n+1},c)-f_{\theta^-}(\hat{x}_{\tau_n},\tau_n,c)\big\|_2^2\Big],
\label{eq:cd}
\end{equation}
where $\hat{x}_{\tau_n}$ is obtained from $x_{\tau_{n+1}}$ by one Euler step of the frozen guided teacher over the grid interval $1/50$. Here $\theta^-$ is the
moving-average copy of the student. The consistency model is sampled by predicting $x_0$, re-noising to the next
level and repeating, but the deployed coarse- and fine-shape students take a single evaluation from pure noise.

\paragraph{Trajectory distillation (structure and texture flows).}
With more than one step, consistency distillation trains for a sampler that re-noises between steps rather than for the deterministic sampler that \sysname runs (the same observation motivates phased consistency models~\citep{rw_wang2024pcm}). Its 4-step structure student indeed agrees less with the 12-step teacher than the trajectory student, and its 4-step texture student agrees less with it than four Euler steps of the released flow (Section~\ref{sec:ablation_distill}, Appendix~\ref{app:ledger}). We therefore distill these two stages directly onto their deployment grid. Let $\mathcal{G}=\{t_{g_0}=1>t_{g_1}>\dots>t_{g_k}=0\}$
be a $(k{+}1)$-point subset of the 13-point grid of the 12-step teacher, and let $x^{\mathrm{T}}_{t}$ denote the state of the teacher's
own deterministic (guided) trajectory at grid time $t$, computed once per training view and cached. The student is
trained to reproduce, with one Euler step, the teacher's multi-step transition between consecutive grid points:
\begin{equation}
\mathcal{L}_{\mathrm{traj}}=\mathbb{E}_{i}\Big[\big\|x^{\mathrm{T}}_{t_{g_i}}+\big(t_{g_{i+1}}-t_{g_i}\big)\,v_\theta\big(x^{\mathrm{T}}_{t_{g_i}},t_{g_i},c\big)-x^{\mathrm{T}}_{t_{g_{i+1}}}\big\|_2^2\Big],
\label{eq:traj}
\end{equation}
with $i$ drawn uniformly over the $k$ segments. The inputs are teacher states (teacher forcing), and re-noising
is never used. At deployment the student is run with $k$ plain Euler steps on $\mathcal{G}$ without guidance. Specifically, the texture student uses $k=2$ with $t=(1,0.75,0)$, two segments of 6 teacher steps. The structure student uses $k=4$ with the non-uniform grid $t=(1,0.9,0.75,0.6,0)$, i.e.,
segments of 3, 3, 2 and 4 teacher steps. We compute the structure trajectories and deploy the student at the time shift 3.0 of the shape and texture flows rather than the 5.0 of the released structure sampler, so its target is the 12-step teacher at this shift (Appendix~\ref{app:distill}).

\paragraph{Relation to prior objectives, and deployment.}
Equation~\eqref{eq:traj} derives from progressive distillation~\citep{rw_salimans2022progressive}: one round at the non-uniform ratios of the structure grid above instead of 2:1, with the guided teacher's ODE states as inputs instead of noised data (further relations: Appendix~\ref{app:distill}). Guidance is folded into the cached teacher states so that the students run conditional-only~\citep{rw_meng2023guided}, as AssetGen does for its shape students~\citep{rw_assetgen2026}. The per-stage choice of objective and step count follows the ablation of Section~\ref{sec:ablations}. At deployment the structure student runs inside a static CUDA graph, which removes launch overhead without changing the arithmetic (counted with the student in Table~\ref{tab:optimizations}). As with the attention kernels, the adapters of the fine-shape and texture students are folded into the bf16 weights ($\approx$), while the structure and coarse-shape students keep the adapter as a separate branch, because bf16 rounding there would change the active voxels (Section~\ref{sec:method-infra-flows}). Since the students do not use guidance, they take 4+1+1+2 = 8 network evaluations per image.

\paragraph{Same-step teachers and rejected alternatives.}
\label{sec:method-steps}
\label{sec:method-sparsity}
To isolate the effect of distillation, we compare every student in Section~\ref{sec:ablation_distill} with its \emph{same-step teacher}, the released flow of its stage sampled with the student's step count and without guidance. Two alternatives to distillation failed (Appendix~\ref{app:ledger}). Shortening the texture flow without training removes only a few evaluations before the quality degrades, because the discretization error grows as the steps drop, and making the steps cheaper fails because the teacher's attention is neither local nor low-rank and little is redundant.

\subsection{Training}
\label{sec:method-training}

\paragraph{Data.}
Under both objectives, we train the students on outputs of the 12-step teachers only, without ground-truth geometry in any loss. The structure student and the two shape students are trained on the \emph{5k set} (5,243 objects), which was built first. The texture student is trained on the \emph{10k set}, 10,000 objects drawn later from TRELLIS-500K~\citep{rw_xiang2025trellis} with a stricter metadata filter and better-framed views. Retraining the structure student and the two shape students on the 10k set did not improve them (Section~\ref{sec:ablations}, Appendix~\ref{app:distill}), and we therefore keep the 5k-set students. Appendix~\ref{app:data} gives the selection funnel, the views of the two sets and the framing check.

\paragraph{Schedules.}
On these sets, we trained the structure student with the trajectory objective for 500 updates on one GPU, the coarse-shape student with the consistency objective for 3,000 updates on 16 GPUs, and the texture student with the trajectory objective for 3,000 updates. The fine-shape student was trained like the coarse-shape student and is the checkpoint at 2,000 updates of a run configured for 3,000. Optimizer settings and costs are in Appendix~\ref{app:distill}, and the selection of every checkpoint in Appendix~\ref{app:protocol}.

\section{Experiments} 
\label{sec:experiments}

\subsection{Setup}
\label{sec:setup}

\paragraph{Data.}
We evaluate on three sets. The main comparison (Section~\ref{sec:comparison}) is on the \emph{Toys4K test part}, 2,653 held-out objects of Toys4K~\citep{ap_toys4k} with one fixed view each, evaluated once after the evaluation code had been frozen. The \emph{Toys4K development part} (500 objects) was evaluated after \sysname was fixed, as a check before the test part (block (b) of Appendix Table~\ref{tab:toys4k_dev}). The \emph{validation set} consists of 400 objects drawn from the same pool as the training data and disjoint from the 10k set (Section~\ref{sec:method-training}). Since \sysname was selected on this set, the numbers measured on it may be subject to selection bias, which is why the main comparison is on the test part. The ablations, the distillation studies, the timing runs, the qualitative figures and the failure analysis are nevertheless all run on it, and its \emph{40-object subset} (two views each, 80 inputs) is used for every timing run and every analysis of Section~\ref{sec:progressive}. Appendix~\ref{app:protocol} describes the splits, the views, the selection and the seven validation objects that also occur in the 5k set.

\paragraph{Baselines.}
We label \sysname as \emph{Ours} in tables and figures and compare it with the released pipeline, with the configurations between the two, and with external baselines. Base (Pixal3D) is the released pipeline run with its own script (Section~\ref{sec:method-overview}), and \emph{Base (Pixal3D), re-run} is the same pipeline and export run inside our measurement code. Two further configurations keep the released 12-step flows and are named after their rows in Table~\ref{tab:strong_baseline}: \emph{mesh path rebuilt} uses the rebuilt mesh path of Section~\ref{sec:method-infra}, and \emph{+ flow optimizations} adds the optimizations of Section~\ref{sec:method-infra-flows} that apply to the released networks. As external baselines, we run TRELLIS.2~\citep{rw_xiang2025trellis2}, TripoSG~\citep{rw_li2025triposg} and Hunyuan3D-2.1~\citep{rw_hunyuan3d21} at their released defaults. We run every method ourselves and evaluate it with the same code, and we do not copy any number from the Pixal3D paper, because its model differs from the released one (Section~\ref{sec:method-overview}).

\paragraph{Metrics.}
The main tables follow Table~1 of the Pixal3D paper: the normal map of each output is rendered in the input-image frame and compared with the ground-truth normal map. We report the IoU of the predicted and ground-truth masks, the PSNR, SSIM~\citep{ap_wang2004ssim} and LPIPS~\citep{ap_lpips}, the mean and median angular error, the mean angular error in a band at the mask boundary ($\mathrm{Mean}_B$) and the accuracy at $11.25^\circ$, $22.5^\circ$ and $30^\circ$. Following the wording of the paper, every metric except the IoU is computed on the overlap of the two masks, the \emph{overlap region}. However, we compute PSNR, SSIM and LPIPS on the full image outside Table~\ref{tab:main} and its validation-set and development-part counterparts unless the overlap region is named. %
As in the paper, we do not align any output to the ground truth: the external baselines are rendered with the ground-truth camera pose, and Base (Pixal3D) and \sysname with their own inference-time projection. The \emph{true-camera} condition gives the model the ground-truth intrinsics, distance and scale instead of the estimated camera and is used only for analysis (Table~\ref{tab:distill}(a), Appendix~\ref{sec:sensitivity}). Throughout, $\Delta$ is the mean of a paired per-object difference with its 95\% bootstrap CI in brackets. ``Significant'' refers to a Holm-adjusted sign test and is marked in tables by $^{\ast}$ (better) and $^{\dagger}$ (worse) against the reference row. Appendix~\ref{app:protocol} gives the remaining choices and the statistics.

\paragraph{End-to-end time and export.}
The E2E time (Section~\ref{sec:method-overview}) includes input preparation and the file write but not model loading, and requests run one at a time on one H200 GPU. Unless a caption states otherwise, speed is the median E2E time per input and quality is measured on the 400 validation objects with the inference-time camera of the released script.
The \emph{official export} is the export setting of Base (Pixal3D) (about 1M faces, $4096^2$ texture). Every result uses it except a few analyses in the appendix that name the \emph{smaller export} (fewer faces and a $2048^2$ texture). Speed ratios thus compare pipelines with the same target face count and texture resolution. However, the assets are not identical: the texture encoding (lossless PNG for ours, lossy WebP for Base (Pixal3D)), the atlas layout and the remesh resolution differ (Section~\ref{sec:method}). Given identical latents, renders of the two exports agree at 32\,dB PSNR (Appendix~\ref{app:profile}). %

We time the configurations above in two runs. The \emph{dedicated timing run} gives the main timing result, with one machine per method and requests strictly one at a time: its medians are 133.0\,s for Base (Pixal3D) and 4.05\,s for \sysname, a ratio of 33$\times$. The cumulative measurement (Section~\ref{sec:method-overview}) instead runs all its configurations in one process, so its rows are paired. We therefore compare times only within a run and take every ratio to Base (Pixal3D) in the dedicated timing run. The E2E column of a table likewise serves only the comparison of its rows (Appendix~\ref{app:profile}).

\subsection{Progressive Design Analysis}
\label{sec:progressive}

\begin{table}[tbp]
\centering
\caption{\textbf{Progressive design analysis.} Base (Pixal3D): dedicated timing run; other rows: the first row and the ends of the three groups of the cumulative measurement. Steps: structure / coarse-shape / fine-shape / texture; critical path: medians of the per-input sums; speed-up: ratio of medians to the row above; $\Delta$: paired difference to Base (Pixal3D).}
\label{tab:strong_baseline}
\tablestyle{3pt}{1.1}
\begin{tabular}{@{}llrrrrrr@{}}
\toprule
 & & & \multicolumn{2}{c}{Critical path (s)} & & \multicolumn{2}{c}{$\Delta$ normal map} \\
\cmidrule(lr){4-5}\cmidrule(l){7-8}
Configuration & Steps & E2E (s) & flows & post-processing & Speed-up & PSNR (full) & Mean ($^\circ$) \\
\midrule
Base (Pixal3D) & 12/12/12/12 & 133.0 & 17.3 & 108.8 & -- & reference & reference \\
\midrule
Base (Pixal3D), re-run & 12/12/12/12 & 138.4 & 17.2 & 118.1 & -- & \makecell[tr]{$-0.03$\\{\scriptsize[$-0.11$, $+0.05$]}} & \makecell[tr]{$+0.13$\\{\scriptsize[$-0.07$, $+0.32$]}} \\
\hspace{0.5em}+ mesh path rebuilt & 12/12/12/12 & 19.0 & 17.2 & 1.32 & \makecell[tr]{$7.26\times$\\{\scriptsize[5.85, 9.00]}} & \makecell[tr]{$+0.09$\\{\scriptsize[$-0.01$, $+0.20$]}} & \makecell[tr]{$+0.03$\\{\scriptsize[$-0.17$, $+0.24$]}} \\
\hspace{0.5em}+ flow optimizations & 12/12/12/12 & 15.4 & 13.8 & 1.32 & \makecell[tr]{$1.24\times$\\{\scriptsize[1.20, 1.27]}} & \makecell[tr]{$+0.09$\\{\scriptsize[$-0.00$, $+0.20$]}} & \makecell[tr]{$+0.04$\\{\scriptsize[$-0.16$, $+0.25$]}} \\
\hspace{0.5em}+ distilled students (\textbf{Ours}) & 4/1/1/2 & 4.00 & 2.38 & 1.35 & \makecell[tr]{$3.84\times$\\{\scriptsize[3.58, 4.31]}} & \makecell[tr]{$+0.44$\\{\scriptsize[$+0.27$, $+0.61$]}} & \makecell[tr]{$-1.47$\\{\scriptsize[$-1.98$, $-1.01$]}} \\
\bottomrule
\end{tabular}
\end{table}

We first attribute the speed-up to the three groups of changes, then apply the rebuilt mesh path to a second pipeline, and finally examine how the quality changes along the cumulative measurement.

\paragraph{The three groups.}
The attribution is based on the cumulative measurement, which starts from the re-run of Base (Pixal3D) inside our measurement code, i.e., from the released pipeline with its own export. The median of this re-run is consistent with that of Base (Pixal3D) in the dedicated timing run (Table~\ref{tab:strong_baseline}), and its time divides between mesh post-processing and the four flows almost as in Base (Pixal3D) (Appendix~\ref{app:profile}). Thus, the speed-up below is not due to the export or to our measurement code. Of the three groups, the first two shorten the time without a measurable change in the input-view normal quality: up to the row ``+ flow optimizations'', every CI of the difference to Base (Pixal3D) includes zero. The third group also changes the quality. Replacing the four flows by the distilled students turns the 15\,s pipeline into a 4\,s one and raises the normal PSNR by 0.34\,dB $[+0.18, +0.50]$ over the 12-step flows of the row above. The mean angular error improves as well, with a CI excluding zero.

\paragraph{Order of attribution.}
The speed-up factors of these three groups (Table~\ref{tab:strong_baseline}) multiply to 34.6$\times$, the ratio from the re-run of Base (Pixal3D) to \sysname in the cumulative measurement. However, because the stages overlap and the bottleneck moves, they are not independent contributions and depend on the order in which the changes are applied. In the reverse order the students and the flow optimizations are applied first on the released, serial post-processing. An estimate from the per-input critical paths then gives only 1.15$\times$ for these two steps together and 30.1$\times$ for the post-processing rebuilt last. In either order, most of the end-to-end factor is due to the mesh post-processing. The split of Base (Pixal3D) is reversed at the end of the measurement: in \sysname the neural flows take 59\% of the mean critical path and the mesh post-processing takes 34\%, mostly the $4096^2$ bake and the file export (dedicated timing run, Appendix Table~\ref{tab:app_profile}).

\paragraph{A second pipeline.}
Next, we test whether the mesh optimizations transfer to a second pipeline. The released TRELLIS.2 pipeline calls the same GLB export code, so we apply our rebuilt mesh path to TRELLIS.2-4B, first without the adaptive half-resolution remesh and then with it (Table~\ref{tab:trellis2_meshpath}).
The median E2E time falls by 6.7$\times$ because almost all of the post-processing leaves the critical path. It still falls by 5.7$\times$ in a conservative estimate that attributes to us none of the slower generation of the official run on some inputs (Appendix~\ref{app:profile}).
At the same time, the normal-map and geometry metrics do not change, which shows that the mesh optimizations are not specific to Pixal3D.
However, two of our choices do not transfer. Running the mesh work concurrently with the texture flow does not save time, because in TRELLIS.2 the texture flow slows down by as much as the mesh work hides. The adaptive half-resolution remesh saves little time at a small cost in base color (Table~\ref{tab:trellis2_meshpath}).

\begin{table}[tbp]
\centering
\caption{\textbf{The rebuilt mesh path in TRELLIS.2.} TRELLIS.2-4B at its released settings on the 40-object subset; only the mesh post-processing is carried over, not the flow engineering (Section~\ref{sec:method-infra-flows}) or the distilled students (Section~\ref{sec:method-model}). Speed-up and $\Delta$: against TRELLIS.2 (official); $\Delta$ color: input-view base-color PSNR against the input image.}
\label{tab:trellis2_meshpath}
\tablestyle{3.5pt}{1.1}
\begin{tabular}{@{}lrrrrrr@{}}
\toprule
 & & & & \multicolumn{2}{c}{$\Delta$ normal map} & $\Delta$ color \\
\cmidrule(lr){5-6}\cmidrule(l){7-7}
Configuration & E2E (s) & Post-processing (s) & Speed-up & PSNR (full) & Mean ($^\circ$) & PSNR \\
\midrule
TRELLIS.2 (official) & 112.7 & 91.4 & -- & reference & reference & reference \\
\quad + rebuilt mesh path, full-resolution remesh & 16.9 & 2.2 & \makecell[tr]{$6.7\times$\\{\scriptsize[5.4, 8.6]}} & \makecell[tr]{$+0.00$\\{\scriptsize[$-0.00$, $+0.01$]}} & \makecell[tr]{$+0.00$\\{\scriptsize[$-0.01$, $+0.02$]}} & \makecell[tr]{$-0.00$\\{\scriptsize[$-0.02$, $+0.01$]}} \\
\quad + adaptive half-resolution remesh & 16.5 & 2.1 & \makecell[tr]{$6.8\times$\\{\scriptsize[5.4, 8.5]}} & \makecell[tr]{$-0.07$\\{\scriptsize[$-0.18$, $+0.00$]}} & \makecell[tr]{$+0.02$\\{\scriptsize[$-0.07$, $+0.11$]}} & \makecell[tr]{$-0.06$\\{\scriptsize[$-0.09$, $-0.04$]}} \\
\bottomrule
\end{tabular}
\end{table}

\paragraph{Quality against time.}
Finally, over the whole cumulative measurement, the input-view normal quality never falls below that of Base (Pixal3D) (Appendix Table~\ref{tab:app_progressive_full}). The Chamfer distance does not change measurably either: its paired difference to Base (Pixal3D) has a CI including zero at every row. The base-color LPIPS improves slightly, as every row from the tiled mesh path on is 0.005--0.006 below Base (Pixal3D), with a CI excluding zero. By comparison, sampling the released flows with fewer steps instead of distilling them preserves the quality down to 8 steps per stage (11.1\,s) and then degrades sharply: at 4 steps (5.3\,s) the pipeline is already slower than \sysname and loses geometry (Appendix Figure~\ref{fig:app_quality_speed_paired}, triangles).

\begin{figure}[tb]
\centering
\includegraphics[width=\linewidth]{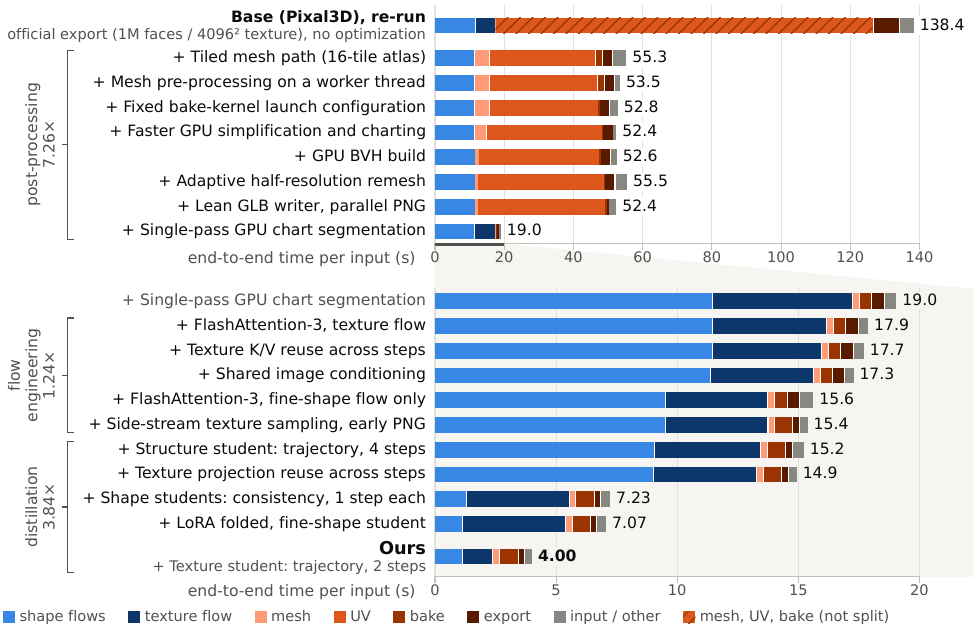}
\caption{\textbf{Cumulative measurement.} Each row adds one change to the row above, grouped as in Section~\ref{sec:method} (left: group factors). Bars: median critical-path time per stage (shape flows: structure, coarse- and fine-shape flows); gray: remainder to the median E2E time; hatched: the released export, timed as one call. Lower panel: row 9 on a 0--20\,s axis.}
\label{fig:waterfall}
\end{figure}

\subsection{Comparison with Baselines}
\label{sec:comparison}

The analysis above measures quality on the validation set only. On the held-out Toys4K test part (2,651 objects with an output from both pipelines), \sysname is significantly better than Base (Pixal3D) on all ten columns of Table~\ref{tab:main} (Appendix Table~\ref{tab:app_paired_test}), and the two pipelines also have the best values in every quality column. The gain is spread over the objects rather than concentrated in a few: normal PSNR on the overlap region rises by 0.70\,dB on average and is higher on five of every six objects. The silhouette IoU is also higher on 59\% of the objects, but a tail of 21 objects on which it is at least 10\,pt below Base (Pixal3D) lowers its mean ($-0.19$\,pt $[-0.29, -0.11]$). On the single-view 3D metrics \sysname is also better than Base (Pixal3D), in Chamfer distance and F-score (Appendix Table~\ref{tab:app_other_test}).

\begin{table}[tbp]
\centering
\caption{\textbf{Single-image generation under the Pixal3D protocol.} The 2,653 objects of the Toys4K test part, evaluated once after the protocol was frozen; input-view normal maps against the ground truth, metrics as in Section~\ref{sec:setup}. E2E: median of the dedicated timing run on the validation set (baselines: Appendix Table~\ref{tab:app_baseline_e2e}).} %
\label{tab:main}
\tablestyle{3.5pt}{1.1}
\begin{tabular}{@{}lccccccccccc@{}}
\toprule
Method & \makecell{IoU\\(pt)\\$\uparrow$} & \makecell{PSNR\\(overlap)\\$\uparrow$} & \makecell{SSIM\\(overlap)\\$\uparrow$} & \makecell{LPIPS\\(overlap)\\$\downarrow$} & \makecell{Mean\\($^\circ$)\\$\downarrow$} & \makecell{Median\\($^\circ$)\\$\downarrow$} & \makecell{Mean$_B$\\($^\circ$)\\$\downarrow$} & \makecell{11.25$^\circ$\\(pt)\\$\uparrow$} & \makecell{22.5$^\circ$\\(pt)\\$\uparrow$} & \makecell{30$^\circ$\\(pt)\\$\uparrow$} & \makecell{E2E (s)\\$\downarrow$} \\
\midrule
TRELLIS.2 & 79.55 & 17.15 & 0.577 & 0.033 & 25.06 & 18.97 & 27.56 & 48.27 & 64.47 & 70.80 & -- \\
TripoSG & 59.87 & 14.22 & 0.504 & 0.051 & 38.87 & 34.31 & 40.88 & 22.74 & 35.89 & 42.98 & -- \\
Hunyuan3D-2.1 & 59.82 & 14.31 & 0.505 & 0.049 & 39.33 & 35.24 & 41.80 & 23.20 & 36.10 & 43.17 & -- \\
\midrule
Base (Pixal3D) & \textbf{96.97} & 20.87 & 0.668 & 0.021 & 14.25 & 9.99 & 15.73 & 61.35 & 82.58 & 88.71 & 133.0 \\
\rowcolor{rowgray}
\textbf{Ours} & 96.78 & \textbf{21.56} & \textbf{0.712} & \textbf{0.018} & \textbf{13.00} & \textbf{9.06} & \textbf{14.33} & \textbf{65.53} & \textbf{85.13} & \textbf{90.40} & \textbf{4.05} \\
\bottomrule
\end{tabular}

\end{table}

The same pattern holds on the Toys4K development part, where all ten differences keep their sign, and on the validation set (Appendix Table~\ref{tab:app_main_validation}). There \sysname is better than Base (Pixal3D) on nine of the ten columns. The tenth is the silhouette IoU, which is on par in the median but lower in the mean due to six failure objects (Section~\ref{sec:failures}).

This advantage comes with three qualifications. The first concerns the alignment scores. On the test part \sysname is slightly higher than Base (Pixal3D) in CLIP~\citep{ap_clip}, but under the point-cloud encoders ULIP-2~\citep{ap_ulip2} and Uni3D~\citep{ap_uni3d} it is slightly and significantly lower, on the test part (Appendix Table~\ref{tab:app_paired}) and on the validation set. These lower scores are caused by the shape rather than by the texture, as the analysis on the validation set shows (Section~\ref{sec:method-infra-mesh}, Appendix~\ref{app:perception_gap}). %
The second concerns the texture, which the protocol of the Pixal3D paper does not measure. To cover it, we compare the unlit base color with the ground-truth base color on the 339 validation objects whose materials can be exported (Appendix Table~\ref{tab:app_appearance_gt}). \Sysname is significantly better than Base (Pixal3D) in SSIM and LPIPS and on par in mean PSNR, but the set of its renders has a significantly larger Fr\'echet distance~\citep{me_stein2023fddino} to the ground-truth renders in DINOv2 features~\citep{ap_dinov2}. Exchanging the texture flow alone shows that this shift is caused by the 2-step texture student, which reproduces each texture of the 12-step teacher with less fine detail (11\% less gradient energy in the renders; Appendix~\ref{app:tables}).
The third concerns the other generators. On the metrics outside the protocol, another generator has the best value in nine of the 14 columns of Appendix Table~\ref{tab:app_other_test}: TRELLIS.2 has the highest F-score at all three thresholds (F@0.01 of 0.523 versus 0.408 for \sysname), and Hunyuan3D-2.1 the highest ULIP-2 and Uni3D scores. For both Pixal3D pipelines, the level of the single-view 3D metrics is determined by the inference-time camera (Section~\ref{sec:ablations}, Appendix~\ref{app:sensitivity}). %

\paragraph{In-the-wild images.} We also ran the second comparison of the Pixal3D paper (its Table~2) on 150 in-the-wild images that we collected (Appendix~\ref{app:tables}, Table~\ref{tab:wild}). As on Toys4K, \sysname is slightly below Base (Pixal3D) on both geometry-only alignment scores, while TRELLIS.2 is above Base (Pixal3D) on all four scores and TripoSG on the two geometry-only ones.

\paragraph{Cost.}
Every method occupies one GPU for the whole request, including the time in which the CPU mesh post-processing of the released pipeline runs, so the GPU-seconds per asset equal the E2E time of the dedicated timing run (Section~\ref{sec:setup}).
The official script does not report the request-time VRAM peak above the resident weights. This peak is 14.2\,GB for \sysname versus 37.2\,GB for the re-run of Base (Pixal3D) in the cumulative measurement (Appendix Table~\ref{tab:app_progressive_full}).

\subsection{Qualitative Results}
\label{sec:qualitative}

\begin{figure}[tbp]
\centering
\includegraphics[width=\linewidth]{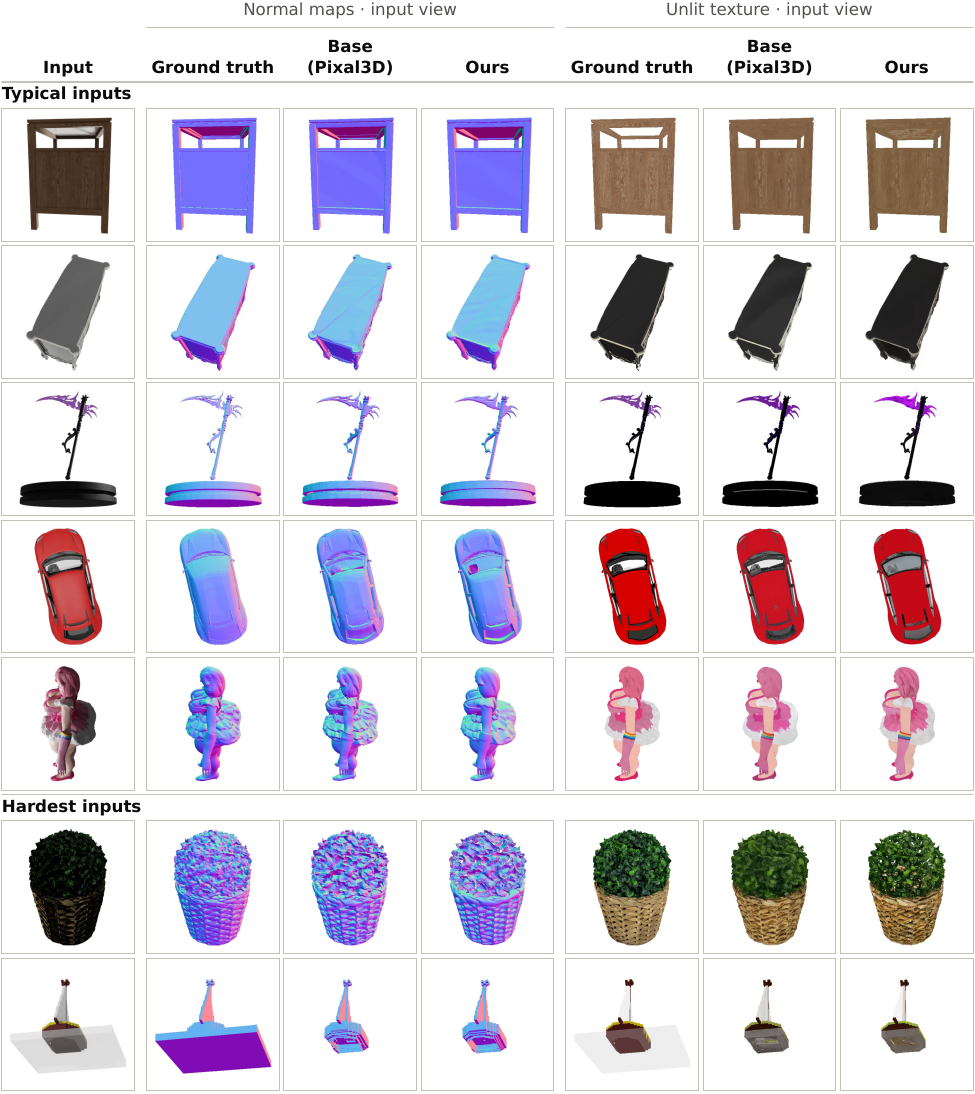}
\caption{\textbf{Qualitative comparison.} Seven inputs of the 40-object subset, rendered from the input camera: the input image, then the normal maps and the unlit base color of the ground truth, Base (Pixal3D) and \emph{Ours}. Rows 1--5: typical inputs, in order of increasing angular error of Base (Pixal3D); rows 6--7: the two inputs with the largest error for \emph{Ours} (selection: Section~\ref{sec:qualitative}).} %
\label{fig:gallery}
\end{figure}

Despite the advantage measured above, the two pipelines are hard to distinguish on single inputs. Figure~\ref{fig:gallery} shows this on seven validation inputs drawn from the 71 inputs of the 40-object subset whose base-color render is not nearly white. Five of them are typical inputs, nearest the 10th, 30th, 50th, 70th and 90th percentiles of the angular error of Base (Pixal3D) among the other 69, one per object. The other two are the inputs with the largest error for \sysname, which are also the two worst for Base (Pixal3D). On the typical rows the two pipelines produce nearly the same geometry and base color, with mean angular errors less than $2.5^\circ$ apart. \Sysname is slightly worse in four of these five rows, but over the 71 inputs the opposite holds. The per-input errors of the two pipelines are almost perfectly rank-correlated, and \sysname is better by more than $1^\circ$ on about half of the inputs and worse by more than $1^\circ$ on only four. %
The failures are also shared: on the potted plant (sixth row) neither pipeline resolves the leaf-level relief of the foliage, and on the voxel sailboat (last row) both omit the translucent slab under the hull. %
The advantage of \sysname in Appendix Table~\ref{tab:app_main_validation} is thus the sum of small differences on many inputs rather than a visible change on single objects.

\paragraph{Failure cases.}
\label{sec:failures}
Beyond the shared failures, Appendix Figure~\ref{fig:failures} shows the six validation objects on which our silhouette IoU is at least 10 pt below that of Base (Pixal3D), and the ablation run of Section~\ref{sec:ablations} traces each to a stage. Three fail only with our structure student and are recovered by the structure student trained on the 10k set, two stem from the 1-step shape students and are recovered by the 2-step shape students, and one is not recovered by any single-stage change. However, the structure student trained on the 10k set again leaves six objects at least 10 pt below Base (Pixal3D). They are two of the original six and four new ones, so the choice of structure student moves these failures between objects without changing their number. %

\begin{table}[tbp]
\centering
\caption{\textbf{One part of \sysname changed at a time.} First block: students replaced by alternatives; second block: each student replaced by its same-step teacher (Section~\ref{sec:method-steps}). Each block is one run, and E2E compares only within a block (\emph{Ours}: first / second run). Second block: $^{\ast}$\,/\,$^{\dagger}$ against the \emph{Ours} row; CD and F@0.01: Appendix~\ref{app:tables}.} %
\label{tab:ablation}
\tablestyle{2.3pt}{1.1}
\begin{tabular}{@{}lccccccccccccc@{}}
\toprule
Method & \makecell{IoU\\(pt)\\$\uparrow$} & \makecell{PSNR\\(full)\\$\uparrow$} & \makecell{SSIM\\(full)\\$\uparrow$} & \makecell{LPIPS\\(full)\\$\downarrow$} & \makecell{Mean\\($^\circ$)\\$\downarrow$} & \makecell{Median\\($^\circ$)\\$\downarrow$} & \makecell{Mean$_B$\\($^\circ$)\\$\downarrow$} & \makecell{11.25$^\circ$\\$\uparrow$} & \makecell{22.5$^\circ$\\$\uparrow$} & \makecell{30$^\circ$\\$\uparrow$} & \makecell{CD\\($\times10^{-3}$)\\$\downarrow$} & \makecell{F@0.01\\$\uparrow$} & \makecell{E2E (s)\\$\downarrow$} \\
\midrule
\rowcolor{rowgray}
\textbf{Ours} & 96.33 & 25.20 & 0.916 & 0.056 & 15.92 & 10.68 & 17.97 & 57.69 & 80.25 & 86.56 & 37.24 & 0.332 & 3.91\,/\,3.79 \\
\midrule
\multicolumn{14}{@{}l}{\textit{Students replaced by alternatives}} \\
Texture student (5k set) & 96.33 & 25.20 & 0.916 & 0.056 & 15.92 & 10.68 & 17.96 & 57.69 & 80.25 & 86.56 & 37.24 & 0.332 & 3.81 \\
2-step shape students & 96.47 & 25.02 & 0.911 & 0.057 & 16.63 & 11.33 & 18.65 & 55.02 & 78.35 & 85.32 & 37.17 & 0.333 & 4.22 \\
Structure student (10k set) & 96.27 & 25.14 & 0.915 & 0.056 & 15.89 & 10.52 & 17.80 & 57.58 & 80.29 & 86.58 & 37.92 & 0.327 & 3.87 \\
\midrule
\multicolumn{14}{@{}l}{\textit{Students replaced by their same-step teachers}} \\
Structure, 4 steps & 93.23$^{\dagger}$ & 24.43$^{\dagger}$ & 0.911$^{\dagger}$ & 0.068$^{\dagger}$ & 17.14 & 11.55$^{\dagger}$ & 18.97$^{\dagger}$ & 55.76$^{\dagger}$ & 78.47 & 84.81 & 38.17$^{\dagger}$ & 0.334 & 3.54 \\
Coarse shape, 1 step & 95.89$^{\dagger}$ & 25.15 & 0.915 & 0.059$^{\dagger}$ & 15.71$^{\ast}$ & 10.54$^{\ast}$ & 17.87 & 58.29$^{\ast}$ & 80.78$^{\ast}$ & 86.94$^{\ast}$ & 37.53$^{\dagger}$ & 0.329$^{\dagger}$ & 3.73 \\
Fine shape, 1 step & 95.98$^{\dagger}$ & 25.01$^{\dagger}$ & 0.916 & 0.058$^{\dagger}$ & 15.98 & 10.66$^{\ast}$ & 18.13$^{\dagger}$ & 58.08$^{\ast}$ & 80.17 & 86.36$^{\dagger}$ & 37.30$^{\dagger}$ & 0.331$^{\dagger}$ & 3.79 \\
\bottomrule
\end{tabular}
\end{table}

\subsection{Ablations}
\label{sec:ablations}

\paragraph{One stage at a time.}
The ablations test the choice of student at each stage. In the first block of Table~\ref{tab:ablation}, replacing the 1-step consistency students of the coarse- and fine-shape flows by 2-step trajectory students is the only change that is significantly worse, in every normal-map column except the silhouette IoU. The configuration is also 0.44\,s slower than \sysname (paired mean, $[+0.38, +0.52]$), so the 1-step students are kept. %
In contrast, the texture student trained on the 5k set and the structure student trained on the 10k set do not differ significantly from \sysname on any metric, including the single-view 3D and the alignment scores (Appendix~\ref{app:tables}). %

\begin{table}[tbp]
\centering
\caption{\textbf{Distillation methods per stage.} (a)~Support IoU against the ground truth and against the 12-step teacher's support. (b)~Agreement with the 12-step teacher (decoded-color L1, relative latent error). Updates: of the student (whole run in parentheses); Stage: texture-stage time; GPU-h: training time $\times$ GPUs (+ teacher-trajectory cache; est.: estimated).} %
\label{tab:distill}
\tablestyle{3pt}{1.1}
\begin{tabular*}{\linewidth}{@{\extracolsep{\fill}}llllllll@{}}
\toprule
\multicolumn{8}{@{}l}{\textbf{(a) Structure stage} (400 views of 278 validation objects, true camera; support IoU at 32$^3$)}\\
\midrule
Method & Data & Steps & IoU$\uparrow$ & \makecell[l]{$\Delta$ vs 12-step\\teacher {[95\% CI]}} & \makecell[l]{IoU vs 12-step\\teacher$\uparrow$} & \makecell[l]{GPU-h\\(+cache)} & Updates \\
\midrule
12-step teacher & --- & 12, guided & 0.597 & ref. & 1.000 & --- & --- \\
Consistency & 10k set & 4 & 0.596 & $-$0.002 [$-$0.007, +0.003] & 0.797 & 27.8 & 12,000 \\
DMD2 & 10k set & 4 & 0.587 & $-$0.010 [$-$0.015, $-$0.006] & 0.785 & 18.0 & 3,000 (15,000) \\
Trajectory (Ours) & 5k set & 4 & 0.598 & +0.001 [$-$0.002, +0.004] & 0.894 & 0.25 (+3.5) & 500 \\
\end{tabular*}

\begin{tabular*}{\linewidth}{@{\extracolsep{\fill}}llllllll@{}}
\midrule
\multicolumn{8}{@{}l}{\textbf{(b) Texture stage} (12 validation objects $\times$ 2 views; agreement with the 12-step teacher)}\\
\midrule
Method & Data & Steps & L1 ($\times10^{-2}$)$\downarrow$ & \makecell[l]{$\Delta$ vs 2-step\\teacher {[95\% CI]}} & Rel. err.$\downarrow$ & Stage (s) & GPU-h \\
\midrule
12-step teacher & --- & 12 & 0.00 & --- & 0.000 & 5.87 & --- \\
2-step teacher & --- & 2 & 3.08 & ref. & 0.316 & 0.98 & --- \\
Trajectory & 5k set & 2 & 3.25 & +0.17 [$-$0.34, +0.95] & 0.235 & 1.28 & 7.0 (+4.0) \\
Trajectory (Ours) & 10k set & 2 & 2.38 & $-$0.70 [$-$0.98, $-$0.44] & 0.200 & 1.28 & 7.0 (+23.0 est.) \\
\bottomrule
\end{tabular*}

\end{table}

\paragraph{Distillation methods per stage.}
\label{sec:ablation_distill}
At the structure stage (Table~\ref{tab:distill}(a)) the support IoU against the ground truth varies little across methods, and only DMD2 is worse than the 12-step teacher with a CI excluding zero ($-0.010$ $[-0.015, -0.006]$). What separates the methods instead is the agreement with the teacher's support, where trajectory distillation leads at a fraction of the training cost of consistency distillation and DMD2. %
In (a) the compared students also differ in training set, but with the sets matched the result is the same (Appendix Table~\ref{tab:distill_ablation_full}).
At the texture stage we compare the students by their agreement with the 12-step teacher, with the same-step teacher as the reference. The student trained on the 5k set matches it, whereas the student trained on the 10k set exceeds it (closer to the 12-step teacher on all 12 objects) at the same texture-stage time, so \sysname uses it. %

\paragraph{Same-step teachers.}
The second block of Table~\ref{tab:ablation} replaces one student of \sysname at a time by its same-step teacher (Section~\ref{sec:method-steps}). The structure student is clearly better than its 4-step teacher, by $3.1$\,pt in silhouette IoU and significantly in most normal-map columns, and the texture student is likewise better than its 2-step teacher (Table~\ref{tab:distill}(b)). The 1-step shape students, however, differ from their teachers in both directions: each is better than its 1-step teacher in silhouette IoU, LPIPS, CD and F@0.01 and worse in some normal-map columns. On balance, we keep the students, whose silhouette IoU and CD are better at every stage and whose alignment scores are equal or better (Appendix~\ref{app:tables}). %

\paragraph{Why one step scores higher.}
In the cumulative measurement, the gain of the distilled pipeline over the 12-step flows on the normal metrics comes almost entirely from the step that replaces the coarse- and fine-shape flows by the 1-step students, which leaves Chamfer distance and F-score (Table~\ref{tab:app_progressive_full}) and silhouette IoU unchanged. The released shape flows sampled in one step score at the same level as the students (second block of Table~\ref{tab:ablation}), indicating that the effect is due to the step count rather than to distillation. Because at $t=1$ the noise carries no information about the sample, one Euler step from pure noise with an exact velocity field returns the mean of the latent given the conditioning~\citep{me_liu2023rectified}. The mean averages out the fine surface variation of the individual samples, and metrics that compare one output with one ground truth pixel by pixel favor such an average, at the cost of distance from the distribution of real samples~\citep{ap_blau2018perception}. Consistent with this, the surfaces are measurably smoother: the mean deviation between the normals of nearby surface points falls by 13\% at this step, and the objects whose roughness falls most are those whose angular error improves most (Appendix~\ref{app:tables}).

\paragraph{Sensitivity and the inference camera.}
The remaining analyses are in the appendix. Appendix~\ref{app:bottleneck} shows that only the structure stage turns a small change into a different sample, and Appendix~\ref{sec:sensitivity} that the angular error is highest on thin and elongated objects and that the E2E time is determined almost entirely by the number of fine-shape tokens. Finally, the inference-time camera rather than the students determines the level of the single-view 3D metrics, which is why the main tables are based on the input-view normal maps (Appendix~\ref{app:tables}).

\section{Conclusion}
\label{sec:conclusion}

In the released image-to-3D pipelines we timed, most of the end-to-end time is spent outside the sampling of the networks. Built on this observation, \sysname writes the textured GLB in 4.05\,s versus 133\,s for the released Pixal3D pipeline, 33$\times$ faster. On the held-out Toys4K test part it is also significantly better on every input-view normal metric of the Pixal3D protocol, at a slightly lower mean silhouette IoU.

Beyond this pipeline, two lessons apply to other cascaded generators. One is to classify every optimization by the stage it acts on and by its effect on the output, and to compare quality on the finished asset rather than on the stage. Here, six of the 18 steps of Table~\ref{tab:optimizations} are output-changing, and these six are the ones a new pipeline has to re-validate.
The other is that the step budget need not be uniform across stages, and that once the flows are short the remaining time is determined by what runs on the critical path rather than by the amount of work. The single change that replaced the CPU UV unwrap by a GPU chart segmentation saved more (30.5\,s) than all kernel changes of the flows together, whereas hiding the GPU mesh work behind the 2-step texture flow gained only 0.03\,s, because the two compete for the same GPU.

\paragraph{Limitations.}
\label{sec:limitations}
The first limitation is the support predicted by the structure stage, which in the tested configurations is the dominant source of error against the ground truth. All single-sample ways of producing it that we tried reach about the same support IoU against the ground truth, whereas replacing it by the ground-truth support while keeping every later stage raises the F-score substantially (Appendix~\ref{app:bottleneck}).
A second limitation is the point-cloud alignment: because its remeshed shells have a thicker inner wall, \sysname scores lower than the released pipeline under the point-cloud embeddings ULIP-2 and Uni3D (Section~\ref{sec:comparison}, Appendix~\ref{app:perception_gap}).
The third is texture detail: the 2-step texture student loses fine detail against the 12-step teacher (Section~\ref{sec:comparison}). Distribution-matching objectives~\citep{rw_yin2024dmd2} for the texture stage are a possible way to recover it, but the texture student that we trained for this purpose by mean-flow distillation~\citep{rw_zhao2026mfd} overshot the fine detail (Appendix~\ref{app:ledger}), which leaves this direction open. %
Finally, the comparison with the released pipeline and the baselines is on the held-out Toys4K test part. Every other measurement is on the validation set, which also served model selection (Section~\ref{sec:setup}).

\bibliographystyle{abbrv}
\bibliography{references} %

\beginappendix

\section{Timing Runs and Profiles}
\label{app:profile}

This section gives the setup of the timing runs of Section~\ref{sec:setup} and the full record behind the decomposition of Section~\ref{sec:progressive}.

\begin{table}[!htb]
\centering
\caption{\textbf{Per-stage profile.} Mean time per input on the critical path; concurrent work listed as hidden work; shares: mean stage time over mean E2E time. Base (Pixal3D), \emph{Ours}: dedicated timing run; re-run, + tiled mesh path: cumulative measurement. UV unwrapping: GPU chart clustering, then CPU xatlas (\emph{Ours}: GPU chart segmentation).}
\label{tab:app_profile}
\tablestyle{4pt}{1.1}
\begin{tabular}{@{}lrrrr@{}}
\toprule
Stage (s, mean) & \makecell{Base\\(Pixal3D)} & \makecell{Base (Pixal3D),\\re-run} & \makecell{+ tiled\\mesh path} & \makecell{Ours} \\
\midrule
Input pre-processing & 0.12 & 0.23 & 0.23 & 0.22 \\
Structure flow & 1.12 & 1.12 & 1.13 & 0.35 \\
Coarse-shape flow & 1.59 & 1.64 & 1.65 & 0.19 \\
Fine-shape flow & 11.76 & 11.54 & 11.47 & 0.71 \\
Texture flow & 6.97 & 7.02 & 3.80 & 1.39 \\
Mesh extraction and remesh & 11.17 & 0.47 & 4.60 & 0.45 \\
UV unwrapping & 94.45 & --- & 30.44 & 0.03 \\
Released export, not split & --- & 145.31 & --- & --- \\
Texture bake & 21.13 & --- & 2.17 & 0.74 \\
GLB export & 6.55 & 7.03 & 2.86 & 0.28 \\
Other (timing residual) & 1.09 & 0.15 & 0.09 & 0.10 \\
\midrule
Hidden work & --- & --- & \makecell[r]{UV 3.29\\+ texture flow 15.72} & \makecell[r]{GPU remesh 0.35\\+ UV 0.62\\+ texture flow 0.04} \\
Share, flows / mesh & 14\% / 85\% & 12\% / 88\% & 31\% / 69\% & 59\% / 34\% \\
\midrule
End-to-end, mean & 155.97 & 174.51 & 58.41 & 4.46 \\
End-to-end, median & 133.02 & 138.37 & 55.33 & 4.05 \\
End-to-end, p10--p90 & 66.0--257.4 & 84.5--283.1 & 37.6--82.0 & 3.11--6.11 \\
\bottomrule
\end{tabular}

\end{table}

\paragraph{Dedicated timing run.}
Each machine of the dedicated timing run has one NVIDIA H200 GPU with 141\,GB and 15 vCPUs of an Intel Xeon host of the Sapphire Rapids generation, with CUDA 12.8 and PyTorch 2.7.0. The first three of the 80 inputs are excluded as warm-up, which leaves 77 inputs.
On the machine of \sysname a fixed reference configuration of ours ran in the same process, alternating in order with \sysname from input to input, and the order changes the pipeline's median by 0.03\,s. The ratio of the medians of Base (Pixal3D) and \sysname, 33$\times$, has a bootstrap 95\% CI of 28--37$\times$ over the inputs, and the median of the per-input ratios is 29.6$\times$ $[25.2, 33.4]$.
These medians exclude model loading, which takes 105\,s for Base (Pixal3D) and, in our process, 105.5\,s for the released networks, 66.5\,s for the shape students and 22.4\,s for the texture student. Base (Pixal3D) runs with TF32 matrix products disabled, as released, and our process runs with them enabled.
To see how much Base (Pixal3D) varies from run to run, we timed a second identical run: its median is 128.4\,s (difference of medians $-4.6$\,s, 95\% CI $[-14.7, +9.1]$), and between the two runs single inputs vary by up to 3$\times$ due to the CPU UV unwrapping.
In Table~\ref{tab:app_profile} the flow rows include conditioning, structure decoding and coarse-to-fine upsampling. UV unwrapping is one unwrap in Base (Pixal3D), part of the released export (not split) in its re-run and 16 unwraps in the tiled mesh path, which take 30.44\,s, of which 2.6\,s run on the GPU. For Base (Pixal3D) the export steps are read from the progress output of the released script (1\,s resolution).

\paragraph{Official and rebuilt export.}
The released script writes a WebP texture and does not expose a face-count or texture-size option. The comparison of the rebuilt mesh path with the released export uses the 74 of the 80 inputs on which both receive identical latents (on the other six the fine-shape token sets differ by one or two voxels). There our export has the same face count (ratio 1.00 $[1.00, 1.01]$) but 0.17$\times$ the UV charts and 0.55$\times$ the seam length. Its atlas occupancy is lower (48.8\% versus 59.5\% of the texels), its median texel density the same (ratio 0.99 $[0.98, 1.02]$) and its file 1.5$\times$ larger. Renders of the two exports agree at 32\,dB PSNR ($2048^2$, four azimuths) with the same mean gradient energy (ratio 1.00 $[0.98, 1.02]$; 1.06 at $4\times$ magnification). %

\begin{figure}[!tb]
\centering
\includegraphics[width=\linewidth]{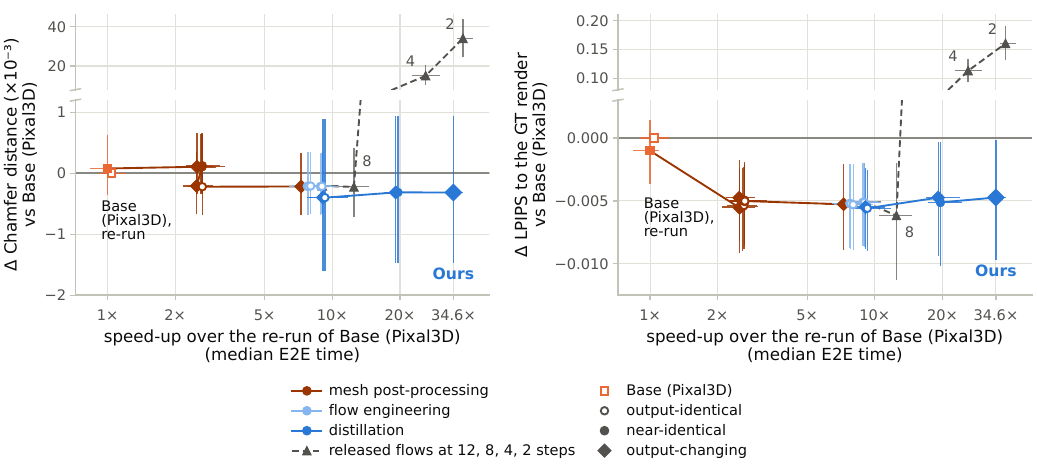}
\caption{\textbf{Quality against speed-up, paired to Base (Pixal3D).} The rows of Figure~\ref{fig:waterfall}: mean paired difference to Base (Pixal3D) in Chamfer distance and base-color LPIPS against the speed-up over the re-run, each with its 95\% CI. Circles: output-identical (open), near-identical (filled); diamonds: output-changing; squares: Base (Pixal3D) (open), its re-run (filled); triangles: the released flows at 12, 8, 4 and 2 steps.}
\label{fig:app_quality_speed_paired}
\end{figure}

\paragraph{Cumulative measurement.}
The classes of Table~\ref{tab:optimizations} are based on the following checks. Of the rows of class =, the two reuse rows were verified to give identical texture latents on all 80 inputs, the shared conditioning and the GLB writer on a few objects, and the bake kernel in a unit test. Mesh post-processing differs from run to run, so the $\approx$ rows are judged against that spread. To check that the cumulative measurement agrees with separate runs, six of its configurations, among them the re-run of Base (Pixal3D), the row ``mesh path rebuilt'' and \sysname, were also measured in separate runs on the same inputs. The ratios of their medians lie between 0.97 and 1.03, each with a 95\% CI that includes 1.
In paired means with 95\% bootstrap CIs over the 40 objects, the mesh post-processing group changes the E2E time by $-150.9$\,s $[-186.8, -122.7]$ from the re-run of Base (Pixal3D), and the flow engineering by a further $-5.34$\,s $[-6.78, -4.16]$. The distilled students, with the texture projection reuse that can only be measured after the structure student (Section~\ref{sec:method-infra-flows}), change it by $-14.0$\,s $[-16.3, -11.9]$, which makes $-170.3$\,s $[-206.0, -142.2]$ in all. Per stage, the flow engineering shortens the fine-shape stage by 2.92\,s, the texture stage by 2.07\,s and the conditioning of the four stages by 0.45\,s, and the distilled students shorten the structure, coarse-shape, fine-shape and texture stages by 0.78, 1.42, 7.89 and 3.95\,s, respectively.
In quality, the Chamfer distance of \sysname relative to the re-run of Base (Pixal3D) does not change significantly ($-0.39\times10^{-3}$) and F@0.01 rises by 0.019 ($[+0.005, +0.035]$).
In Table~\ref{tab:app_progressive_full}, the $\Delta$VRAM of 56.5\,GB in the five rows from the texture K/V reuse to the structure student is a transient peak of the K/V reuse.
The 95\% CI of the median of its first row, the re-run of Base (Pixal3D), is $[126.3, 165.1]$\,s and contains the median of Base (Pixal3D) in the dedicated timing run. The measurement runs on eight machines, each with all configurations in one process, and with the rebuilt mesh path the texture flow is the longer branch on all 80 inputs. In every table on the 400 validation objects other than Table~\ref{tab:app_main_validation}, the E2E column is the median in the run of that table or table block. For reference, over the same 80 inputs the medians of \sysname in this measurement and in the dedicated timing run differ by 0.03\,s.
The base-color LPIPS of Figure~\ref{fig:app_quality_speed_paired} is measured on the 39 objects with a ground-truth base color. Its gain is not due to the texture encoding: re-encoding the textures of the tiled-mesh-path row as lossy WebP, as in the official export, moves its LPIPS by only $+0.0011$, so that $-0.0034$ [$-0.0052$, $-0.0018$] against the re-run remains for the tiled mesh path itself (tiled extraction, atlas and baking). The released flows are sampled with fewer steps in the last row of the flow-engineering group. At 2 steps the pipeline loses more geometry than at 4, and with one step the mesh extraction runs out of GPU memory.

\paragraph{One-at-a-time removal.}
With the students in place, we measured the contribution of FlashAttention-3 in the texture flow, the texture K/V reuse and the texture projection reuse by removing them one at a time (one separate run on one machine). The removals change the E2E time of \sysname by $+0.129$, $+0.006$ and $-0.008$\,s at the median of the paired per-input differences, and only the first of the three CIs excludes zero. The sampling loop of the texture flow, without conditioning and decoding, takes 0.86\,s at the median in \sysname, and switching the worker thread of the mesh pre-processing off makes it 0.147\,s faster while lengthening the E2E time by $+0.034$\,s $[+0.027, +0.059]$.

\begin{table}[!tb]
\centering
\caption{\textbf{Every step of the cumulative measurement.} $\Delta$ Mean, $\Delta$ PSNR: paired mean difference to Base (Pixal3D) in the mean angular error and the normal PSNR on the full image, CI beneath; first row: the re-run. CD and F@0.01: Appendix~\ref{app:tables}. $\Delta$VRAM: request-time peak above the resident weights, maximum over inputs.}
\label{tab:app_progressive_full}
\tablestyle{1.0pt}{1.1}
\begin{tabular}{@{}lcccccccccc@{}}
\toprule
 & & & & & & \multicolumn{2}{c}{E2E (s)$\downarrow$} & & & \\
\cmidrule(lr){7-8}
Configuration & \makecell{PSNR$\uparrow$\\(full)} & \makecell{SSIM$\uparrow$\\(full)} & \makecell{Mean$\downarrow$\\($^\circ$)} & \makecell{CD$\downarrow$\\($\times10^{-3}$)} & F@0.01$\uparrow$ & median & p90 & \makecell{$\Delta$VRAM\\(GB)$\downarrow$} & $\Delta$ Mean$\downarrow$ & $\Delta$ PSNR$\uparrow$ \\
\midrule
Base (Pixal3D), re-run & 24.27 & 0.9048 & 17.49 & 36.37 & 0.305 & 138.4 & 283.1 & 37.2 & \makecell[tr]{$+0.13$\\{\scriptsize[$-0.07$, $+0.32$]}} & \makecell[tr]{$-0.03$\\{\scriptsize[$-0.11$, $+0.05$]}} \\
+ tiled mesh path (16-tile atlas) & 24.27 & 0.9048 & 17.47 & 36.40 & 0.304 & 55.3 & 82.0 & 29.1 & \makecell[tr]{$+0.10$\\{\scriptsize[$-0.08$, $+0.28$]}} & \makecell[tr]{$-0.03$\\{\scriptsize[$-0.10$, $+0.04$]}} \\
+ mesh pre-processing on a worker thread & 24.28 & 0.9048 & 17.46 & 36.39 & 0.304 & 53.5 & 85.2 & 30.9 & \makecell[tr]{$+0.10$\\{\scriptsize[$-0.08$, $+0.28$]}} & \makecell[tr]{$-0.03$\\{\scriptsize[$-0.10$, $+0.04$]}} \\
+ fixed bake-kernel launch configuration & 24.28 & 0.9048 & 17.46 & 36.40 & 0.304 & 52.8 & 81.8 & 31.3 & \makecell[tr]{$+0.10$\\{\scriptsize[$-0.08$, $+0.29$]}} & \makecell[tr]{$-0.03$\\{\scriptsize[$-0.10$, $+0.05$]}} \\
+ faster GPU simplification and charting & 24.28 & 0.9048 & 17.46 & 36.42 & 0.304 & 52.4 & 81.4 & 31.1 & \makecell[tr]{$+0.10$\\{\scriptsize[$-0.08$, $+0.28$]}} & \makecell[tr]{$-0.03$\\{\scriptsize[$-0.10$, $+0.04$]}} \\
+ GPU BVH build & 24.28 & 0.9048 & 17.46 & 36.40 & 0.304 & 52.6 & 76.2 & 31.0 & \makecell[tr]{$+0.10$\\{\scriptsize[$-0.08$, $+0.29$]}} & \makecell[tr]{$-0.03$\\{\scriptsize[$-0.10$, $+0.05$]}} \\
+ adaptive half-resolution remesh & 24.40 & 0.9048 & 17.40 & 36.08 & 0.309 & 55.5 & 79.3 & 26.7 & \makecell[tr]{$+0.04$\\{\scriptsize[$-0.16$, $+0.25$]}} & \makecell[tr]{$+0.09$\\{\scriptsize[$-0.01$, $+0.20$]}} \\
+ lean GLB writer, parallel PNG & 24.40 & 0.9047 & 17.40 & 36.07 & 0.310 & 52.4 & 80.2 & 26.7 & \makecell[tr]{$+0.04$\\{\scriptsize[$-0.16$, $+0.24$]}} & \makecell[tr]{$+0.09$\\{\scriptsize[$-0.01$, $+0.19$]}} \\
+ single-pass GPU chart segmentation & 24.40 & 0.9048 & 17.40 & 36.08 & 0.310 & 19.0 & 41.9 & 26.8 & \makecell[tr]{$+0.03$\\{\scriptsize[$-0.17$, $+0.24$]}} & \makecell[tr]{$+0.09$\\{\scriptsize[$-0.01$, $+0.20$]}} \\
+ FlashAttention-3, texture flow & 24.40 & 0.9048 & 17.40 & 36.08 & 0.310 & 17.9 & 38.3 & 26.8 & \makecell[tr]{$+0.04$\\{\scriptsize[$-0.16$, $+0.24$]}} & \makecell[tr]{$+0.09$\\{\scriptsize[$-0.01$, $+0.20$]}} \\
+ texture K/V reuse across steps & 24.40 & 0.9047 & 17.40 & 36.08 & 0.310 & 17.7 & 37.9 & 56.5 & \makecell[tr]{$+0.04$\\{\scriptsize[$-0.16$, $+0.24$]}} & \makecell[tr]{$+0.09$\\{\scriptsize[$-0.01$, $+0.20$]}} \\
+ shared image conditioning & 24.40 & 0.9048 & 17.40 & 36.08 & 0.310 & 17.3 & 37.6 & 56.5 & \makecell[tr]{$+0.04$\\{\scriptsize[$-0.17$, $+0.24$]}} & \makecell[tr]{$+0.09$\\{\scriptsize[$-0.01$, $+0.20$]}} \\
+ FlashAttention-3, fine-shape flow only & 24.40 & 0.9047 & 17.40 & 36.08 & 0.310 & 15.6 & 31.2 & 56.5 & \makecell[tr]{$+0.04$\\{\scriptsize[$-0.16$, $+0.24$]}} & \makecell[tr]{$+0.10$\\{\scriptsize[$-0.01$, $+0.20$]}} \\
+ side-stream texture sampling, early PNG & 24.40 & 0.9048 & 17.40 & 36.08 & 0.310 & 15.4 & 31.3 & 56.5 & \makecell[tr]{$+0.04$\\{\scriptsize[$-0.16$, $+0.25$]}} & \makecell[tr]{$+0.09$\\{\scriptsize[$-0.00$, $+0.20$]}} \\
+ structure student: trajectory, 4 steps & 24.40 & 0.9036 & 17.32 & 35.90 & 0.324 & 15.2 & 28.9 & 56.5 & \makecell[tr]{$-0.04$\\{\scriptsize[$-0.26$, $+0.18$]}} & \makecell[tr]{$+0.09$\\{\scriptsize[$-0.03$, $+0.22$]}} \\
+ texture projection reuse across steps & 24.40 & 0.9036 & 17.32 & 35.89 & 0.324 & 14.9 & 28.9 & 24.7 & \makecell[tr]{$-0.04$\\{\scriptsize[$-0.26$, $+0.18$]}} & \makecell[tr]{$+0.09$\\{\scriptsize[$-0.03$, $+0.22$]}} \\
+ shape students: consistency, 1 step each & 24.76 & 0.9139 & 15.93 & 35.98 & 0.324 & 7.23 & 13.0 & 14.2 & \makecell[tr]{$-1.43$\\{\scriptsize[$-1.93$, $-0.98$]}} & \makecell[tr]{$+0.45$\\{\scriptsize[$+0.29$, $+0.63$]}} \\
+ LoRA folded, fine-shape student & 24.74 & 0.9144 & 15.89 & 35.98 & 0.324 & 7.07 & 12.8 & 14.2 & \makecell[tr]{$-1.47$\\{\scriptsize[$-1.98$, $-1.01$]}} & \makecell[tr]{$+0.44$\\{\scriptsize[$+0.27$, $+0.61$]}} \\
\textbf{Ours} (+ texture student: trajectory, 2 steps) & 24.74 & 0.9144 & 15.89 & 35.98 & 0.324 & 4.00 & 5.91 & 14.2 & \makecell[tr]{$-1.47$\\{\scriptsize[$-1.98$, $-1.01$]}} & \makecell[tr]{$+0.44$\\{\scriptsize[$+0.27$, $+0.61$]}} \\
\bottomrule
\end{tabular}

\end{table}

\begin{table}[!tb]
\centering
\caption{\textbf{End-to-end time of the released pipelines.} The 77 inputs of the dedicated timing run; shares: means of per-input shares. Sampling: conditioning and flow or diffusion sampling; post-processing: the rest, including latent decoding. $^{a}$\,Dedicated timing run. $^{b}$\,Timed while generating the baseline outputs. $^{c}$\,Sampling includes multi-view diffusion and super-resolution.}
\label{tab:app_baseline_e2e}
\tablestyle{5pt}{1.1}
\begin{tabular}{@{}llrrrrr@{}}
\toprule
 & & \multicolumn{3}{c}{End-to-end time (s)} & \multicolumn{2}{c}{Median seconds (mean share)} \\
\cmidrule(lr){3-5}\cmidrule(lr){6-7}
Method & Output & median & mean & p90 & sampling & post-processing \\
\midrule
Base (Pixal3D)$^{a}$ & 1M faces, $4096^2$ texture & 133.0 & 156.0 & 257.4 & 17.3 (16\%) & 109.3 (83\%) \\
TRELLIS.2$^{b}$ & 1M faces, $4096^2$ texture & 105.3 & 128.5 & 197.7 & 14.3 (16\%) & 85.8 (84\%) \\
Hunyuan3D-2.1$^{b,c}$ & 40k faces, $2048^2$ texture & 62.2 & 64.1 & 81.8 & 13.5 (22\%) & 48.4 (78\%) \\
TripoSG$^{b}$ & 1.2M faces, no texture & 7.9 & 7.8 & 10.6 & 2.7 (37\%) & 4.9 (60\%) \\
\bottomrule
\end{tabular}

\end{table}

\paragraph{The other released pipelines.}
Table~\ref{tab:app_baseline_e2e} times the three baselines on the 77 inputs of the dedicated timing run at their released defaults, in the run that generated their outputs, on several machines of the same type. Latent decoding, itself a network, runs in one call with the iso-surface extraction and is counted as post-processing. The TRELLIS.2 median here and the 112.7\,s of Table~\ref{tab:trellis2_meshpath} come from different runs. Sampling is the smaller part in all four pipelines, but what follows differs. Mesh processing and export take 83\% of the time in Base (Pixal3D) and TRELLIS.2 and 62\% in Hunyuan3D-2.1, but 9\% in TripoSG, where field decoding and iso-surface extraction take 51\%. For TripoSG the statement that the networks are not the main cost therefore holds for the sampling of its flow only. For Base (Pixal3D), summed from the progress output of the released script instead of timed around the export call, the post-processing is 108.8\,s (Table~\ref{tab:strong_baseline}).

\paragraph{The rebuilt mesh path in TRELLIS.2.} In Table~\ref{tab:trellis2_meshpath} the generation stages run the same code in both configurations and their per-input times agree, but on some inputs the official run generated more slowly. A conservative estimate that attributes none of this difference to us (the generation times of the official run plus our post-processing) still gives 5.7$\times$. The base-color renders of the two outputs agree about as closely as two exports of the same latents through our path (42.7 versus 43.1\,dB PSNR).

\paragraph{Timings reported by related works.} In FlashVDM the volume decoding takes longer than the diffusion sampling~\citep{rw_lai2025flashvdm}, and in SAM~3D the mesh decoder is the slowest module~\citep{rw_feng2026fastsam3d}, but neither profile reports UV unwrapping, baking or export separately. SF3D~\citep{rw_boss2024sf3d} reports its total and the 150\,ms of its export, and Meshy T2~\citep{rw_xu2026meshyt2} its total for an untextured mesh. AssetGen~\citep{rw_assetgen2026} times simplification, hidden-face removal, parallelized xatlas unwrapping (2.6\,s) and GPU normal baking. In its ablation of the shape branch, GPU geometry processing saves 25\,s before distillation, and of the 41\,s saved after distillation, 20\,s come from the attention and compilation kernels, 16\,s from parallel UV unwrapping and GPU normal baking and 5\,s from overlapped I/O. Its distillation takes 100\,$\to$\,25 and 20\,$\to$\,5 steps for shape and 32\,$\to$\,16\,$\to$\,8\,$\to$\,4 for texture.

\paragraph{Chart segmentation parameters.}
The parameters of the chart segmentation of Section~\ref{sec:method-infra-mesh} are as follows. Faces take the nearest of 26 fixed directions (the cube axes and the edge and corner diagonals). Charts of at most 32 faces merge into the neighbor with the longest shared boundary if their mean normals are within 45$^\circ$, charts of more than 1,500 faces are cut in half, and charts of fewer than 8 faces join a neighbor within 70$^\circ$. Each chart is projected orthographically along its area-weighted mean normal at the best of 16 in-plane rotations, with its border vertices duplicated to avoid a solver, and packed with a 2-texel gutter into rows of decreasing height that fill 90\% of the atlas. A chart that covers a texel twice in the $4096^2$ rasterization is split into depth layers, for at most 6 passes. The charts are found as connected components with a GPU union-find, and everything except the packing loop runs on the GPU on a worker thread.

\paragraph{Released pipeline.} The model of the Pixal3D paper conditions the geometry on DINOv2 features~\citep{ap_dinov2} back-projected into a 3D feature volume, with only the global DINOv2 token passed through cross-attention. In the released implementation, in each transformer block the DINOv3 features sampled where a token's voxel projects into the input image are added to the output of a cross-attention over the global (class and register) tokens of the encoder. In the profile of Section~\ref{sec:method-overview}, the $4096^2$ texture bake comprises attribute sampling, inpainting and material assembly, and Table~\ref{tab:strong_baseline} reports medians where Table~\ref{tab:app_profile} reports means.

\paragraph{Mesh path details.} In the tiled mesh path, large charts are cut into pieces of bounded area, the 16 groups have equal face counts, and the xatlas instances run on a thread pool. The path also builds the BVH in parallel and replaces the CPU inpainting that pads the texture by a GPU pull-push fill. Its 16-tile CPU unwrapping hides the 12-step texture flow on most inputs, which is why rows 2--8 of Figure~\ref{fig:waterfall} lack a texture-flow segment. Figure~\ref{fig:atlas_paths} shows the tiled mesh path with the latents of the released flows and \sysname with those of the distilled students. The adaptive half-resolution remesh halves all 80 inputs of the 40-object subset. In a control run in one process on the 80 inputs, adding it to the row ``+ single-pass GPU chart segmentation'' of Table~\ref{tab:app_progressive_full} changes the mean E2E time by $-0.83$\,s [$-1.54$, $-0.42$].

\section{Data and Evaluation Protocol}
\label{app:protocol}

This section defines the training sets of Section~\ref{sec:method-training}, the evaluation sets and statistics of Section~\ref{sec:setup}, and how the configuration and the checkpoints were selected.

\paragraph{10k set and validation set.}\label{app:data}
Both come from the Sketchfab (Objaverse~\citep{ap_deitke2023objaverse}), 3D-FUTURE~\citep{ap_fu2021future3d}, HSSD~\citep{ap_khanna2024hssd} and ABO~\citep{ap_collins2022abo} parts of TRELLIS-500K~\citep{rw_xiang2025trellis} (188,934 objects). Metadata-only rules, applied in this order, remove duplicates of Toys4K evaluation objects (25 objects, flagged by the same file hash, a Toys4K token in the name, tags or description, the same shortest and longest captions, or a caption TF-IDF cosine $\geq 0.85$) and broken files (3). They then remove objects without a caption (669), scenes (20,526), objects with surroundings (5,964), multi-object sets (2,294), flat or text items (281), scans (229), game-world exports (161) and extreme face counts (1,971), and keep 156,811 objects (83.0\%). We did not add an aesthetic threshold to that of TRELLIS-500K or search for geometric near-duplicates.
From this pool, stratified by source and face-count bin, we draw the 10k set (10,000 objects, two random views each, rendered with the released renderer), its matched subset for the data ablation (5,243 objects, as many as the 5k set) and, disjoint from it, the 400 validation objects. One object of the 10k set lacks structure-support data and is absent from all training caches, so training uses 19,998 views of 9,999 objects.

\paragraph{5k set.}
The 5k set consists of 3,743 objects from Sketchfab and 1,500 from HSSD, with two fixed views each. An automatic framing check flags 57.1\% of its original conditioning images and 0.6\% of the re-rendered views, and 35.4\% of its captions describe scenes, not single objects. The consistency-distilled students saw one original conditioning image per object. The texture student trained on the 5k set used 4,986 objects, and the structure student of \sysname two controlled re-rendered views of each of its 5,243 objects (10,486 views).

\paragraph{Toys4K.}
TRELLIS-500K lists 3,229 objects of Toys4K~\citep{ap_toys4k} (aesthetic score $\geq 4.5$). We set aside a list of 105 Toys4K objects for checks during development, and 76 of them are among the 3,229, which leaves 3,153 objects. These are split once, stratified by category, into a 500-object development part and a 2,653-object test part. Base (Pixal3D) produced no mesh for two of the 2,653 test objects and TripoSG for one, so means are taken over the objects with an output and paired differences over the 2,651 objects with both. Every method receives the same fixed view of each object (yaw 45$^\circ$, pitch 25$^\circ$, field of view 40$^\circ$, camera distance 2.532 in the units of the released renderer).

\paragraph{Subsets of the validation set.}
Each validation object is rendered from four random views with the released renderer, and quality is evaluated on one of them per object, fixed in advance. The conditioning image and the ground-truth camera of a validation input come from the same render. The 40-object subset is a random draw with a fixed seed, stratified by data source. Two per-stage ablations use other samples (Table~\ref{tab:distill}). The 400 views of the structure stage are a fixed random sample of the rendered views and cover 278 objects.

\paragraph{Validation objects in the 5k set.}
Seven validation objects occur in the 5k set. Without them the comparison on the validation set does not change. On the remaining 393 objects the paired differences between \sysname and Base (Pixal3D) in normal PSNR on the overlap region, mean angular error, silhouette IoU and ULIP-2 are each within 0.01 of the 400-object values, and every Holm-adjusted result on the validation set is unchanged. Two results elsewhere cross the significance threshold. The CMMD difference of Table~\ref{tab:app_appearance_gt} becomes significant on the 332 appearance objects outside the 5k set ($+0.0072$, Holm $p$ 0.063 $\to$ 0.028, CI still including zero). In the structure-stage ablation (271 objects) the lower score of DMD2 with the Euler sampler against the 12-step teacher (Table~\ref{tab:distill_ablation_full}) loses its Holm significance ($p$ 0.037 $\to$ 0.087). The exchange of the texture flow in Table~\ref{tab:app_appearance_gt} keeps all nine of its results.

\paragraph{Metric details.} Since the Pixal3D paper does not specify the resolution, the background, the LPIPS network or the boundary band, we fix them to $512^2$, a white background, AlexNet and a 10-px band. For Base (Pixal3D) on the Toys4K test part, for example, the overlap-region and full-image versions give a PSNR of 20.87 and 28.86\,dB (Tables~\ref{tab:main} and~\ref{tab:app_main_full}). The external baselines are rendered with the ground-truth camera pose after a fixed map of their output frame. In a table, bold marks the best value among the compared rows, PSNR is in dB and angular errors are in degrees.

\paragraph{Statistics.}
Paired comparisons use objects as the unit (the 2,651 paired objects of the Toys4K test part, 400 objects of the validation set, or the 40 objects of the subset with their two views averaged). Confidence intervals are 95\% percentile intervals of 10,000 bootstrap resamples of the objects. For per-input times of the 40-object subset the two inputs of an object are resampled together. ``Significant'' means an exact two-sided sign test on the per-object differences, Holm-adjusted at $p<0.05$ within one of three families. These are the 13 normal-map metrics (the ten columns of Table~\ref{tab:main}, with PSNR, SSIM and LPIPS on both the overlap region and the full image), the 4 single-view 3D metrics and the 7 image--shape alignment metrics. Two tables use other families: Table~\ref{tab:app_appearance_gt} (18 tests fixed in advance: its nine metrics at two export settings, of which the table shows the official one) and Table~\ref{tab:distill_ablation_full} (the 18 runs of block (A)). The sign test counts objects, so it can be significant although the confidence interval of the mean difference includes zero. The rows compared in a table or table panel are computed with one version of the evaluation code. Exceptions: the Effect column of Table~\ref{tab:optimizations} and the step effects quoted from it resample inputs, and FD and CMMD use 1,000 resamples and a permutation test with 1,000 permutations.

\paragraph{Selection of the configuration and the checkpoints.}
The three alternatives of Table~\ref{tab:ablation} were compared with \sysname on the 400 validation objects by the tests above at the official export, and on the 40-object subset with a visual review of its inputs. Every candidate returned an asset for each of the 400 inputs. In time, the structure student trained on the 10k set is not distinguishable (paired mean $+0.02$\,s $[-0.06, +0.11]$), and the texture student trained on the 5k set is 0.03\,s faster at the median of the paired differences (mean $-0.01$\,s $[-0.09, +0.08]$). The texture student trained on the 10k set was chosen because it is closer to the 12-step teacher on 11 of the 12 texture objects (views averaged per object; exact sign test $p = 0.006$; L1 values: Table~\ref{tab:distill}(b)).
The 4-step structure student is the checkpoint after 500 updates, which is 0.003 better in support IoU against the ground truth than the one after 250 ($[+0.001, +0.006]$) on the 400 views of the structure stage. The coarse-shape student was trained until a stopping rule fixed in advance was met, evaluated on the 105 Toys4K objects set aside above. The choice of the fine-shape checkpoint (Section~\ref{sec:method-training}) was not recorded, and the texture student is the last checkpoint of its run.

\section{Full Results}
\label{app:tables}

\begin{figure}[!tb]
\centering
\includegraphics[trim=21pt 0pt 2pt 0pt,clip,width=\linewidth,height=0.87\textheight,keepaspectratio]{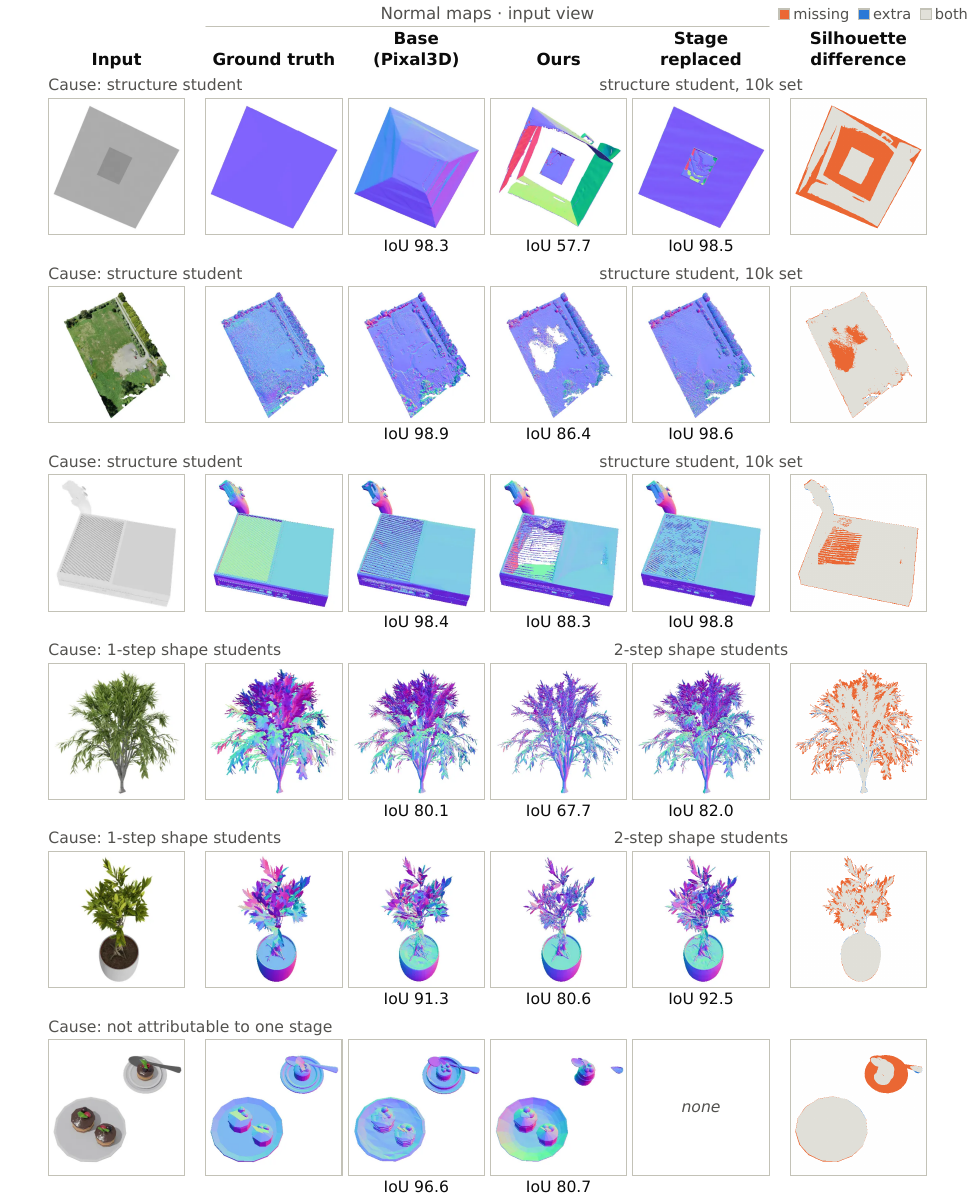} %
\caption{\textbf{The six failure objects of Section~\ref{sec:failures}.} Per row: the input; normal maps of the ground truth, Base (Pixal3D), \emph{Ours} and \emph{Ours} with one stage replaced (rows 1--3: structure student trained on the 10k set; rows 4--5: 2-step shape students; row 6: none recovers it), each with its silhouette IoU; silhouette difference to the ground truth (orange: missing; blue: extra).}
\label{fig:failures}
\end{figure}

This section gives the full versions of the comparisons of Sections~\ref{sec:comparison} and~\ref{sec:ablations}, and the failure objects of Section~\ref{sec:failures} (Figure~\ref{fig:failures}).

\begin{table}[!tb]
\centering
\caption{\textbf{Table~\ref{tab:main} on the validation set and on the Toys4K development part.} (a)~The 400 validation objects, which also served model selection, one view each; E2E as in Table~\ref{tab:main}. (b)~The 500 development objects, one fixed view each (Appendix~\ref{app:protocol}); E2E not measured on this set.}
\label{tab:app_main_validation}\label{tab:toys4k_dev}
\tablestyle{3.5pt}{1.1}
\begin{tabular}{@{}lccccccccccc@{}}
\toprule
Method & \makecell{IoU\\(pt)\\$\uparrow$} & \makecell{PSNR\\(overlap)\\$\uparrow$} & \makecell{SSIM\\(overlap)\\$\uparrow$} & \makecell{LPIPS\\(overlap)\\$\downarrow$} & \makecell{Mean\\($^\circ$)\\$\downarrow$} & \makecell{Median\\($^\circ$)\\$\downarrow$} & \makecell{Mean$_B$\\($^\circ$)\\$\downarrow$} & \makecell{11.25$^\circ$\\(pt)\\$\uparrow$} & \makecell{22.5$^\circ$\\(pt)\\$\uparrow$} & \makecell{30$^\circ$\\(pt)\\$\uparrow$} & \makecell{E2E (s)\\$\downarrow$} \\
\midrule
\multicolumn{12}{@{}l}{\textit{(a) Validation set (400 objects, one view each)}} \\
TRELLIS.2 & 71.88 & 15.80 & 0.521 & 0.084 & 28.92 & 21.07 & 34.00 & 42.13 & 56.97 & 64.18 & -- \\
TripoSG & 60.35 & 13.39 & 0.477 & 0.105 & 40.45 & 34.57 & 45.59 & 17.18 & 34.95 & 45.35 & -- \\
Hunyuan3D-2.1 & 64.90 & 14.47 & 0.499 & 0.093 & 34.56 & 27.87 & 40.75 & 28.47 & 45.11 & 54.56 & -- \\
\midrule
Base (Pixal3D) & \textbf{96.60} & 19.20 & 0.613 & 0.053 & 17.22 & 11.68 & 19.42 & 53.36 & 77.32 & 84.63 & 133.0 \\
\rowcolor{rowgray}
\textbf{Ours} & 96.33 & \textbf{19.82} & \textbf{0.655} & \textbf{0.048} & \textbf{15.92} & \textbf{10.68} & \textbf{17.96} & \textbf{57.69} & \textbf{80.25} & \textbf{86.55} & \textbf{4.05} \\
\midrule
\multicolumn{12}{@{}l}{\textit{(b) Toys4K development part (500 objects, one fixed view each)}} \\
TRELLIS.2 & 80.70 & 17.44 & 0.587 & 0.031 & 23.70 & 17.39 & 26.12 & 50.37 & 66.92 & 72.92 & -- \\
TripoSG & 58.59 & 14.04 & 0.506 & 0.050 & 39.15 & 34.65 & 40.59 & 21.90 & 34.74 & 41.69 & -- \\
Hunyuan3D-2.1 & 58.56 & 14.04 & 0.505 & 0.048 & 40.05 & 35.83 & 41.92 & 21.67 & 34.03 & 41.05 & -- \\
\midrule
Base (Pixal3D) & \textbf{96.89} & 20.79 & 0.666 & 0.021 & 14.40 & 10.03 & 15.83 & 61.55 & 82.47 & 88.62 & -- \\
\rowcolor{rowgray}
\textbf{Ours} & 96.59 & \textbf{21.47} & \textbf{0.712} & \textbf{0.019} & \textbf{13.16} & \textbf{9.10} & \textbf{14.42} & \textbf{65.72} & \textbf{85.21} & \textbf{90.38} & -- \\
\bottomrule
\end{tabular}

\end{table}

\begin{table}[!tbp]
\centering
\caption{\textbf{Other metrics of Table~\ref{tab:main}.} The Toys4K test part. Top: normal map on the full image and single-view 3D metrics; bottom: image--shape alignment with textured renders (CLIP), normal-map renders (CLIP-N) and point clouds with (col.) or without (geo.) vertex colors (ULIP-2, Uni3D). Definitions: paragraph ``Metrics beyond the protocol''. TripoSG has no texture. $^{\ast}$\,/\,$^{\dagger}$: sign test on the 2,651 objects with both outputs (families: Appendix~\ref{app:protocol}).}
\label{tab:app_3d}\label{tab:app_main_full}\label{tab:app_other_test}
\tablestyle{3.5pt}{1.1}
\begin{tabular}{@{}lccccccc@{}}
\toprule
 & \multicolumn{3}{c}{Normal map} & \multicolumn{4}{c}{Single-view 3D} \\
\cmidrule(lr){2-4}\cmidrule(lr){5-8}
Method & \makecell{PSNR$\uparrow$\\(full)} & \makecell{SSIM$\uparrow$\\(full)} & \makecell{LPIPS$\downarrow$\\(full)} & \makecell{CD$\downarrow$\\($\times10^{-3}$)} & F@0.005$\uparrow$ & F@0.01$\uparrow$ & F@0.02$\uparrow$ \\
\midrule
TRELLIS.2 & 23.15 & 0.929 & 0.076 & 28.01 & 0.302 & 0.523 & 0.712 \\
TripoSG & 19.52 & 0.908 & 0.159 & 58.62 & 0.124 & 0.239 & 0.381 \\
Hunyuan3D-2.1 & 19.76 & 0.910 & 0.158 & 59.94 & 0.143 & 0.264 & 0.400 \\
\midrule
Base (Pixal3D) & 28.86 & 0.956 & 0.024 & 26.61 & 0.207 & 0.392 & 0.598 \\
\rowcolor{rowgray}
\textbf{Ours} & 29.27$^{\ast}$ & 0.961$^{\ast}$ & 0.022$^{\ast}$ & 25.55$^{\ast}$ & 0.220$^{\ast}$ & 0.408$^{\ast}$ & 0.614$^{\ast}$ \\
\bottomrule
\end{tabular}
\vspace{4pt}
\tablestyle{3.5pt}{1.1}
\begin{tabular}{@{}lccccccc@{}}
\toprule
 & \multicolumn{7}{c}{Image--shape alignment} \\
\cmidrule(lr){2-8}
Method & CLIP$\uparrow$ & \makecell{CLIP$\uparrow$\\(input)} & CLIP-N$\uparrow$ & \makecell{ULIP-2$\uparrow$\\(col.)} & \makecell{ULIP-2$\uparrow$\\(geo.)} & \makecell{Uni3D$\uparrow$\\(col.)} & \makecell{Uni3D$\uparrow$\\(geo.)} \\
\midrule
TRELLIS.2 & 0.881 & 0.889 & 0.797 & 0.424 & 0.373 & 0.394 & 0.367 \\
TripoSG & -- & -- & 0.779 & 0.397 & 0.397 & 0.373 & 0.373 \\
Hunyuan3D-2.1 & 0.846 & 0.852 & 0.779 & 0.442 & 0.397 & 0.403 & 0.378 \\
\midrule
Base (Pixal3D) & 0.874 & 0.894 & 0.796 & 0.423 & 0.372 & 0.391 & 0.364 \\
\rowcolor{rowgray}
\textbf{Ours} & 0.877$^{\ast}$ & 0.896$^{\ast}$ & 0.796 & 0.415$^{\dagger}$ & 0.365$^{\dagger}$ & 0.385$^{\dagger}$ & 0.357$^{\dagger}$ \\
\bottomrule
\end{tabular}

\end{table}

\paragraph{Metrics beyond the protocol.}
Besides the normal maps, Table~\ref{tab:app_3d} reports single-view 3D metrics, with every mesh normalized by its own bounding box (CD: symmetric mean L2 distance over 100k surface points), and image--shape alignment. The alignment is the cosine similarity of the input image with textured renders from four yaw views (CLIP; input view only: CLIP (input)), with normal-map renders (CLIP-N) and with a point-cloud embedding of the mesh with vertex colors (col.) or geometry only (geo.) (ULIP-2, Uni3D). Since TripoSG outputs untextured meshes, it has no CLIP score and equal col.\ and geo.\ scores.

\paragraph{In-the-wild images.} The second comparison of the Pixal3D paper (its Table~2) scores in-the-wild images without ground truth by the alignment between the input image and the output, the cosine similarity of a point-cloud embedding and an image embedding with ULIP-2 and Uni3D. We ran this comparison with the frozen evaluation code on 150 images that we collected, Internet photographs with recorded licenses and generated images, three of which show more than one object. Since the paper's images are different, the scores are not comparable with its table (Table~\ref{tab:wild}). On the geometry-only scores, the differences between \sysname and Base (Pixal3D) have CIs excluding zero (Uni3D $-0.74$ $[-1.06, -0.44]$ on the $\times 100$ scale).

\paragraph{Same-step teachers.}
The results of the first block of Table~\ref{tab:ablation} do not change on the overlap region. For the structure stage the same-step teacher is sampled on the student's grid. Because the structure and coarse-shape students run their adapter as a separate branch (Section~\ref{sec:method-model}), the same-step teachers are as fast as or faster than the students at their stage. The remaining metrics are as follows. The structure student is also better in F@0.02 and in three of the seven alignment scores (ULIP-2 colored and geometry only, Uni3D geometry only), and its PSNR, SSIM and LPIPS on the overlap region do not differ significantly. In F@0.005 and F@0.01 the result is mixed, as the teacher's mean F is slightly higher (CI including zero) while \sysname is better on 226 and 224 of the 400 objects (significant by the sign test), so the F@0.01 cell of Table~\ref{tab:ablation} is left unmarked. With the teacher, moreover, one object loses almost its whole shape and, without a boundary band, is left out of the Mean$_B$ of that row, which favors the teacher. The coarse-shape student is also better in LPIPS on the overlap region, F@0.02 and three alignment scores (CLIP-N, ULIP-2 and Uni3D geometry only) and worse in PSNR ($-0.13$\,dB) and SSIM ($-0.0052$) on the overlap region and in mean angular error ($+0.21^\circ$ $[+0.13, +0.29]$). The fine-shape student is also better in PSNR and LPIPS on the overlap region (its silhouette IoU is better on 384 of 400 objects) and worse in SSIM on the overlap region ($-0.0083$ $[-0.0114, -0.0056]$). Its alignment scores do not differ significantly.
Of the single-view 3D and alignment metrics of the run of the first block of Table~\ref{tab:ablation}, only CD and F@0.01 are tabulated. On these metrics, the largest difference between \sysname and the texture student trained on the 5k set or the structure student trained on the 10k set is CD $+0.69\times10^{-3}$ for the structure student (Holm $p = 0.20$). The 2-step shape students, in contrast, have a slightly but significantly lower CD ($-0.07\times10^{-3}$, Holm $p = 0.043$) and significantly lower ULIP-2 (geometry only, $-0.003$) and Uni3D (colored, $-0.003$) scores. Since the 2-step shape students run with the texture student trained on the 5k set, their E2E time in Table~\ref{tab:ablation} should be compared with the row above it. For the structure student trained on the 10k set, the CIs of the mean differences in CD and F-score exclude zero.

\begin{table}[!tb]
\centering
\caption{\textbf{Paired statistics of Table~\ref{tab:main}.} The 2,651 Toys4K test objects with an output of both pipelines. \emph{Ours} $-$ Base (Pixal3D): mean difference with 95\% CI, objects better / worse / tied, sign-test $p$ Holm-adjusted within each block; $^{\ast}$\,/\,$^{\dagger}$: significantly better / worse.}
\label{tab:app_paired}\label{tab:app_paired_test}
\tablestyle{5pt}{1.1}
\begin{tabular}{@{}lrrcr@{}}
\toprule
Metric & $\Delta$ (Ours $-$ Base (Pixal3D)) & 95\% CI & better / worse / tie & $p_\mathrm{Holm}$ \\
\midrule
\multicolumn{5}{l}{\textit{Input-view normal maps (Holm over 13)}} \\
IoU (pt)$^{\ast}$ & $-0.19$ & [$-0.29$, $-0.11$] & 1555 / 1095 / 1 & $3.9\times10^{-19}$ \\
PSNR (overlap)$^{\ast}$ & $+0.70$ & [$+0.66$, $+0.73$] & 2212 / 439 / 0 & $2.7\times10^{-282}$ \\
SSIM (overlap)$^{\ast}$ & $+0.0442$ & [$+0.0421$, $+0.0462$] & 2268 / 383 / 0 & $<10^{-300}$ \\
LPIPS (overlap)$^{\ast}$ & $-0.0025$ & [$-0.0028$, $-0.0022$] & 1975 / 676 / 0 & $9.4\times10^{-146}$ \\
PSNR (full)$^{\ast}$ & $+0.42$ & [$+0.38$, $+0.45$] & 1956 / 695 / 0 & $3.5\times10^{-137}$ \\
SSIM (full)$^{\ast}$ & $+0.0046$ & [$+0.0043$, $+0.0049$] & 2126 / 525 / 0 & $3.2\times10^{-226}$ \\
LPIPS (full)$^{\ast}$ & $-0.0023$ & [$-0.0026$, $-0.0019$] & 1902 / 749 / 0 & $2.7\times10^{-114}$ \\
Mean ($^\circ$)$^{\ast}$ & $-1.25$ & [$-1.32$, $-1.19$] & 2273 / 378 / 0 & $<10^{-300}$ \\
Median ($^\circ$)$^{\ast}$ & $-0.94$ & [$-1.00$, $-0.87$] & 2160 / 491 / 0 & $2.0\times10^{-247}$ \\
Mean$_B$ ($^\circ$)$^{\ast}$ & $-1.41$ & [$-1.48$, $-1.33$] & 2253 / 398 / 0 & $4.6\times10^{-312}$ \\
11.25$^\circ$ (pt)$^{\ast}$ & $+4.22$ & [$+3.99$, $+4.45$] & 2193 / 457 / 1 & $1.2\times10^{-269}$ \\
22.5$^\circ$ (pt)$^{\ast}$ & $+2.55$ & [$+2.39$, $+2.71$] & 2096 / 555 / 0 & $1.6\times10^{-208}$ \\
30$^\circ$ (pt)$^{\ast}$ & $+1.69$ & [$+1.57$, $+1.82$] & 2052 / 599 / 0 & $3.8\times10^{-184}$ \\
\midrule
\multicolumn{5}{l}{\textit{Single-view 3D geometry (Holm over 4)}} \\
CD ($\times 10^{-3}$)$^{\ast}$ & $-1.08$ & [$-1.38$, $-0.79$] & 1577 / 1074 / 0 & $2.8\times10^{-22}$ \\
F@0.005$^{\ast}$ & $+0.0128$ & [$+0.0103$, $+0.0153$] & 1648 / 999 / 4 & $4.0\times10^{-36}$ \\
F@0.01$^{\ast}$ & $+0.0159$ & [$+0.0123$, $+0.0196$] & 1602 / 1043 / 6 & $3.8\times10^{-27}$ \\
F@0.02$^{\ast}$ & $+0.0159$ & [$+0.0116$, $+0.0201$] & 1515 / 1107 / 29 & $1.7\times10^{-15}$ \\
\midrule
\multicolumn{5}{l}{\textit{Image--shape alignment (Holm over 7)}} \\
CLIP, four views$^{\ast}$ & $+0.0028$ & [$+0.0019$, $+0.0036$] & 1541 / 1110 / 0 & $2.3\times10^{-16}$ \\
CLIP, input view$^{\ast}$ & $+0.0017$ & [$+0.0007$, $+0.0027$] & 1424 / 1227 / 0 & $2.8\times10^{-4}$ \\
CLIP-N, four views & $+0.0001$ & [$-0.0005$, $+0.0007$] & 1348 / 1303 / 0 & 0.39 \\
ULIP-2 (col.)$^{\dagger}$ & $-0.0088$ & [$-0.0101$, $-0.0076$] & 1064 / 1587 / 0 & $1.3\times10^{-23}$ \\
ULIP-2 (geo.)$^{\dagger}$ & $-0.0071$ & [$-0.0082$, $-0.0059$] & 1144 / 1507 / 0 & $5.7\times10^{-12}$ \\
Uni3D (col.)$^{\dagger}$ & $-0.0052$ & [$-0.0061$, $-0.0044$] & 1022 / 1629 / 0 & $1.8\times10^{-31}$ \\
Uni3D (geo.)$^{\dagger}$ & $-0.0069$ & [$-0.0077$, $-0.0060$] & 958 / 1693 / 0 & $7.6\times10^{-46}$ \\
\bottomrule
\end{tabular}

\end{table}

\begin{table}[!tb]
\centering
\caption{\textbf{In-the-wild images, protocol of the Pixal3D paper's Table~2.} Image--shape alignment ($\times 100$) with ULIP-2 and Uni3D, geometry only and colored, on 150 images we collected (100 Internet photographs, 50 generated), scored on the 149 with an output from every method. No ground truth; descriptive. TripoSG has no texture.}
\label{tab:wild}
\tablestyle{3.5pt}{1.1}
\begin{tabular}{@{}lcccc@{}}
\toprule
 & \multicolumn{2}{c}{Geometry only} & \multicolumn{2}{c}{Colored} \\
\cmidrule(lr){2-3}\cmidrule(lr){4-5}
Method & Uni3D$\uparrow$ & ULIP-2$\uparrow$ & Uni3D$\uparrow$ & ULIP-2$\uparrow$ \\
\midrule
TRELLIS.2 & 30.70 & 31.38 & 33.34 & 36.67 \\
TripoSG & 30.42 & 31.43 & 30.42 & 31.43 \\
Hunyuan3D-2.1 & 26.88 & 26.88 & 29.88 & 32.46 \\
\midrule
Base (Pixal3D) & \textbf{29.88} & \textbf{30.23} & \textbf{32.88} & \textbf{35.64} \\
\rowcolor{rowgray}
\textbf{Ours} & 29.14 & 29.17 & 32.58 & 34.94 \\
\bottomrule
\end{tabular}

\end{table}

\begin{table}[!tb]
\centering
\caption{\textbf{Shape after the same similarity ICP for every method.} Shape against the ground truth in its unit sphere; object means; $\Delta$: \emph{Ours} $-$ Base (Pixal3D). Metrics: paragraph ``Shape after alignment''.}
\label{tab:app_shape_icp}
\tablestyle{3pt}{1.1}
\begin{tabular*}{\linewidth}{@{\extracolsep{\fill}}lccccc@{}}
\toprule
Method & CD-$\ell_2$ ($\times10^{-3}$)$\downarrow$ & CD-$\ell_1$ ($\times10^{-2}$)$\downarrow$ & EMD ($\times10^{-2}$)$\downarrow$ & F@0.02$\uparrow$ & F@0.05$\uparrow$ \\
\midrule
\multicolumn{6}{@{}l}{\textit{(a) Validation set (400 objects, one view each)}} \\
\midrule
TRELLIS.2 & 1.87 & 2.85 & 9.92 & 0.521 & 0.878 \\
TripoSG & 3.89 & 4.10 & 12.53 & 0.383 & 0.761 \\
Hunyuan3D-2.1 & 2.11 & 3.03 & 10.59 & 0.500 & 0.858 \\
\cmidrule(l){1-6}
Base (Pixal3D) & 3.65 & 3.77 & 11.59 & 0.447 & 0.796 \\
\textbf{Ours} & 3.42 & 3.67 & 11.34 & 0.455 & 0.803 \\
\quad $\Delta$ & \makecell[tc]{$-0.23$\\{\scriptsize [$-0.52$, $-0.03$]}} & \makecell[tc]{$-0.09$\\{\scriptsize [$-0.19$, $-0.02$]}} & \makecell[tc]{$-0.26$\\{\scriptsize [$-0.48$, $-0.05$]}} & \makecell[tc]{$+0.007$\\{\scriptsize [$+0.002$, $+0.013$]}} & \makecell[tc]{$+0.007$\\{\scriptsize [$+0.002$, $+0.013$]}} \\
\midrule
\multicolumn{6}{@{}l}{\textit{(b) Toys4K development part (500 objects, one fixed view each)}} \\
\midrule
TRELLIS.2 & 0.97 & 2.15 & 7.59 & 0.637 & 0.939 \\
TripoSG & 1.62 & 2.74 & 9.06 & 0.525 & 0.886 \\
Hunyuan3D-2.1 & 1.34 & 2.46 & 8.39 & 0.573 & 0.912 \\
\cmidrule(l){1-6}
Base (Pixal3D) & 1.31 & 2.44 & 8.12 & 0.573 & 0.916 \\
\textbf{Ours} & 1.28 & 2.42 & 8.00 & 0.578 & 0.918 \\
\quad $\Delta$ & \makecell[tc]{$-0.03$\\{\scriptsize [$-0.09$, $+0.05$]}} & \makecell[tc]{$-0.02$\\{\scriptsize [$-0.06$, $+0.01$]}} & \makecell[tc]{$-0.12$\\{\scriptsize [$-0.25$, $+0.02$]}} & \makecell[tc]{$+0.006$\\{\scriptsize [$+0.001$, $+0.011$]}} & \makecell[tc]{$+0.002$\\{\scriptsize [$-0.002$, $+0.006$]}} \\
\bottomrule
\end{tabular*}

\end{table}

\paragraph{Shape after alignment.}
A separate comparison measures the shape alone. Every method, Base (Pixal3D) and \sysname included, is aligned to the ground-truth surface by the same similarity ICP (rotation, translation and uniform scale; symmetric nearest-neighbor correspondences on 10,000 surface points per side; 24 axis-aligned starting rotations, of which the one with the smallest symmetric Chamfer distance is kept). Prediction and ground truth are then scaled together so that the ground truth lies in the unit sphere. We report CD-$\ell_2$ and CD-$\ell_1$ (half the sum of the mean squared, or mean, nearest-neighbor distances in the two directions, 10,000 points per side), the EMD (mean distance under an exact one-to-one assignment of 2,048 points per side) and the F-score at 0.02 and 0.05. Only the 3D surfaces enter the alignment. In Table~\ref{tab:app_shape_icp}, TRELLIS.2 ranks first on both sets, and against Base (Pixal3D) \sysname is better in all five columns on the validation set but only in F@0.02 on the Toys4K development part.

\paragraph{Surface roughness.}
The roughness of Section~\ref{sec:ablations} is the mean deviation between the normals of surface points less than 0.01--0.02 apart (about four to seven times the median edge length). At the step that adds the 1-step shape students it falls to a ratio of $0.87$ $[0.82, 0.93]$ at $0.01$, and the fall in roughness correlates with the improvement in angular error across objects (Spearman $\rho = 0.51$). It is measured on the 80 inputs of the cumulative measurement and is lower on 33 of the 40 objects at this step, at an unchanged face count and bounding box. At the scale of one or two faces the roughness does not change, and the structure-student step changes neither the roughness nor the metrics. Within this step, the contributions of the coarse- and fine-shape stages are not separated.

\begin{table}[!tb]
\centering
\caption{\textbf{Appearance against the ground-truth base color.} Unlit base color of the 339 validation objects with exportable materials; metrics: paragraph ``Appearance against the ground truth''. $^{\ast}$\,/\,$^{\dagger}$: Holm-adjusted over a family of 18 tests fixed in advance (Appendix~\ref{app:protocol}). Last rows: the texture flow exchanged alone, in one separate run.}
\label{tab:app_appearance_gt}
\tablestyle{2.5pt}{1.1}
\begin{tabular}{@{}lccccccccc@{}}
\toprule
 & \multicolumn{3}{c}{Input view} & \multicolumn{3}{c}{All views} & \multicolumn{3}{c}{Distance to GT renders} \\
\cmidrule(lr){2-4}\cmidrule(lr){5-7}\cmidrule(lr){8-10}
Method & PSNR$\uparrow$ & SSIM$\uparrow$ & LPIPS$\downarrow$ & PSNR$\uparrow$ & SSIM$\uparrow$ & LPIPS$\downarrow$ & \makecell{FD$\downarrow$\\(Inception)} & \makecell{FD$\downarrow$\\(DINOv2)} & CMMD$\downarrow$ \\
\midrule
Base (Pixal3D) & 24.34 & 0.915 & 0.066 & 16.50 & 0.829 & 0.233 & 32.37 & 292.8 & 0.0557 \\
\rowcolor{rowgray}
\textbf{Ours} & 24.30 & 0.920$^{\ast}$ & 0.061$^{\ast}$ & 16.58$^{\ast}$ & 0.835$^{\ast}$ & 0.225$^{\ast}$ & 32.68 & 313.0$^{\dagger}$ & 0.0624 \\
\midrule
12-step teacher & 24.21 & 0.920 & 0.061 & 16.59 & 0.834 & 0.226 & 31.75 & 293.5 & 0.0421 \\
Difference & $+0.10$ & 0.000 & 0.000 & $-0.01$ & 0.000 & 0.000 & $+0.94$ & $+18.7$ & $+0.0202$ \\
\quad 95\% CI & \makecell{\scriptsize[$-0.05$,\\ \scriptsize$+0.20$]} & \makecell{\scriptsize[$-0.001$,\\ \scriptsize$+0.001$]} & \makecell{\scriptsize[$-0.001$,\\ \scriptsize$+0.001$]} & \makecell{\scriptsize[$-0.13$,\\ \scriptsize$+0.06$]} & \makecell{\scriptsize[$-0.001$,\\ \scriptsize$+0.001$]} & \makecell{\scriptsize[$-0.001$,\\ \scriptsize 0.000]} & \makecell{\scriptsize[$+0.39$,\\ \scriptsize$+1.74$]} & \makecell{\scriptsize[$+14.1$,\\ \scriptsize$+26.4$]} & \makecell{\scriptsize[$+0.0154$,\\ \scriptsize$+0.0257$]} \\
\bottomrule
\end{tabular}

\end{table}

\paragraph{Appearance against the ground truth.}
We compare the unlit base color of each output with the ground-truth base color exported from the asset's own PBR materials (Table~\ref{tab:app_appearance_gt}). Per object we report PSNR, SSIM and LPIPS at the input view under the inference-time projection and over four ground-truth camera views, where the output is placed with its own bounding-box normalization. Per method we report three distances between the set of its renders (these five views per object) and the set of ground-truth renders: the Fr\'echet distance (FD) with Inception features~\citep{me_heusel2017fid} and with DINOv2 features~\citep{me_stein2023fddino,ap_dinov2}, and CMMD~\citep{me_jayasumana2024cmmd}. The 61 objects whose materials the export does not support are left out for every method. On them our advantage in the normal metrics is, if anything, larger than on the others, so the selection does not favor \sysname.
Over all views and at the input view \sysname is better than Base (Pixal3D) in SSIM ($+0.006$ $[+0.004, +0.008]$ over all views) and LPIPS ($-0.007$ $[-0.011, -0.004]$). Its mean PSNR is on par ($+0.08$\,dB $[-0.23, +0.36]$ over all views, higher on 202 of the 339 objects, hence the mark). The mean is dominated by seven objects with a uniform white base color, which every method reproduces almost exactly and on which our baked texture is one gray level below white, an offset of the export path rather than of the texture student. Without these seven objects our PSNR is higher by 0.24\,dB $[+0.13, +0.35]$. The set distances are worse for \sysname: FD (DINOv2) rises by 20.2 $[+7.6, +35.1]$ (significant), CMMD rises as well (Holm $p = 0.063$ in the 18-test family of the table), and FD (Inception) does not change. To see whether the texture student accounts for this, we exchanged the texture flow alone (last rows of Table~\ref{tab:app_appearance_gt}): the exchange leaves the per-object means over all views within 0.01\,dB and 0.001 but raises all three set distances. With the 12-step teacher in place of the texture student, our pipeline is at the level of Base (Pixal3D) in FD (DINOv2) and below it in CMMD. The 2-step texture student loses fine detail rather than color. Its renders have 11\% less gradient energy and 10\% less Laplacian variance than those of the 12-step teacher, which are themselves below the ground-truth renders in both, while the mean color changes by about one gray level and the saturation not at all.

\section{Additional Analyses}
\label{app:diagnostics}

This section gives three analyses behind Sections~\ref{sec:comparison} and~\ref{sec:conclusion}: how the results depend on the object and on the inference camera, how much the structure stage limits the shape, and why the point-cloud embeddings favor Base (Pixal3D).

\subsection{Sensitivity to Objects and the Inference Camera}
\label{app:sensitivity}\label{sec:sensitivity}

\paragraph{Objects.}
On the 400 validation objects the mean angular error is highest for objects with many thin parts and for elongated objects (Spearman $\rho = +0.34$ with the thin-voxel share of the ground-truth support and $-0.33$ with its elongation). The Chamfer distance rises with the thin-voxel share ($\rho = +0.10$) but not with the elongation ($+0.08$, CI including zero), and the elongated objects do not have a lower normal PSNR or F-score. The E2E time is determined almost entirely by the number of fine-shape tokens ($\rho = 0.98$).
The thin-voxel share counts the voxels of the $32^3$ ground-truth support with at most six occupied 26-neighbors, and the elongation is $\sqrt{\lambda_{\min}/\lambda_{\max}}$, small when elongated. These correlations use the released inference path, the smaller export and the texture student trained on the 5k set, which leaves the geometry unchanged.

\paragraph{Inference camera.}
On the 40-object subset we also generated the outputs with the true camera in place of the released inference path. With the released inference path the input-view normal metrics change little and in both directions (PSNR $+0.28$\,dB, $[-0.48, +0.86]$; mean angular error $+0.78^\circ$, $[-0.14, +1.92]$). However, the Chamfer distance is 2.2 times higher ($36.0$ versus $16.3\times10^{-3}$) and F@0.01 falls from 0.566 to 0.324, because the estimated camera scatters the depth proportions per object (absolute log depth/width error 0.167 versus 0.075). The gap remains after the best similarity alignment ($+7.2\times10^{-3}$) and is largest for views from below and for depth-dominant objects.
The released 12-step flows without any of our optimizations show the same gap in the same comparison ($+19.4$ versus $+19.7\times10^{-3}$ for \sysname), so the gap is a property of the released inference path rather than of the students. The error of the estimated field of view does not explain it (Spearman $\rho = 0.11$ with the per-input gap). The true camera is an oracle: it passes the object's extent along the viewing axis through the ground-truth scale.

\subsection{The Structure Stage Sets the Ceiling}
\label{app:bottleneck}

At the structure stage a new noise seed alone moves the support of the 12-step teacher to an IoU of 0.81 against its previous sample, and the nine distilled students lie in the same band (0.79--0.89 against the teacher's support; Table~\ref{tab:distill_ablation_full}). The IoU against the ground truth nevertheless changes by at most 0.006 against the 12-step teacher (0.597), except without guidance ($-0.013$), for DMD2 ($-0.010$; $-0.012$ with the Euler sampler) and for the consistency student with the Euler sampler ($-0.010$). All sixteen generators of the support (the teacher at seven settings and the nine distilled students of Table~\ref{tab:distill_ablation_full}) therefore lie within 0.585--0.600. Only selection among four teacher samples leaves this band: an oracle that keeps the best sample by the ground truth reaches 0.623, and a 2-of-4 vote without the ground truth 0.610 at four times the cost of the stage.
To measure what the support costs downstream, we replace the support that enters the coarse-shape stage of \sysname (40-object subset, true camera, smaller export). The support of the structure student trained on the 10k set (IoU 0.76 with ours) changes nothing (F@0.01 0.566 $\to$ 0.578, CI includes zero). In the full pipeline on the 400 validation objects at the smaller export, the same replacement changes the Chamfer distance of 15\% of the objects by more than $5\times10^{-3}$ in either direction with a mean shift whose CI includes zero. Replacing the 1-step shape students by the 2-step shape students, in contrast, does so on only 0.5\% of the objects but lowers the normal PSNR systematically ($-0.16$ dB, $[-0.20, -0.12]$). The ground-truth support raises the support IoU to 0.918 (not 1, because it is built from a different discretization than the evaluator's) and F@0.01 from 0.566 to 0.959, and lowers CD from 16.3 to 4.3 ($\times 10^{-3}$). Fixing this one stage thus removes 91\% of the F-score shortfall and 73\% of the CD. How the remaining error divides between the coarse-shape, fine-shape and texture stages is not measured.

\subsection{Why the Point-Cloud Embeddings Favor Base (Pixal3D)}
\label{app:perception_gap}

\paragraph{Cause.}
ULIP-2 and Uni3D sample their point clouds uniformly by area from the exported mesh, including never-visible surfaces, unlike the metrics of Table~\ref{tab:main}. Both pipelines export double-walled shells (an outer and an inner sheet two grid voxels apart, from the $\epsilon$-offset of an unsigned distance field~\citep{rw_hou2023dcudf}) with about 1.6 times the ground-truth surface area, so roughly half of the sampled points lie on the invisible inner wall.
Along the cumulative measurement on the 40-object subset, the step that adds the adaptive half-resolution remesh lowers both scores, the step that adds the trajectory-distilled structure student changes neither, and in a per-step evaluation no other step lowers either score by more than 0.002. The gap is thus due to this one change. The re-run of Base (Pixal3D) itself is 0.005 below Base (Pixal3D) in Uni3D ($[-0.010, -0.001]$) and level in ULIP-2 ($+0.001$).
The half-resolution grid doubles the wall, and both embeddings respond to the wall thickness monotonically. With the walls of the sampled points matched, the difference between \sysname and Base (Pixal3D) disappears or reverses, on 60 random validation objects and on the 40-object subset, where Uni3D goes from $-0.012$ to $-0.002$ ($[-0.006, +0.002]$) and ULIP-2 from $-0.009$ to $+0.008$ ($[+0.001, +0.016]$).
Other candidate causes are excluded. The gap is as large for ULIP-2 and Uni3D on the geometry only (Table~\ref{tab:app_paired}, test part). On the same 60 objects, exchanging the colors of the point clouds changes ULIP-2 by $+0.0002$ (exchanging the shapes: $-0.014$), and removing small disconnected components changes either score by at most 0.003. On the surface hidden from the input view our F@0.01 against the ground truth is not lower ($+0.029$), and the six failure objects of Section~\ref{sec:failures} contribute less than 0.0003 to the mean.

\paragraph{Remedy tested.}
Halving the distance band of the remesh, which would bring the wall on the half-resolution grid back to the released thickness, instead merges the two sheets into one non-manifold sheet. The embeddings rise above Base (Pixal3D), but the normal PSNR drops by 2.3\,dB and the mean angular error rises by $9.1^\circ$, so we keep the adaptive half-resolution remesh.

\section{Distillation Runs and Training}
\label{app:distill}

\begin{table}[!tbp]
\centering
\caption{\textbf{All structure-stage distillation runs evaluated on the validation set.} Columns as in Table~\ref{tab:distill}(a), plus precision (P), recall (R), thin-part recall (Thin R) and IoU against the teacher's support (vs T). Data: training views in parentheses. $\blacktriangledown$\,/\,$\blacktriangle$: worse / better than the 12-step teacher, CI excluding zero. GPU-h (GPUs); n/r: not recorded.}
\label{tab:distill_ablation_full}
\tablestyle{3pt}{1.1}
\begin{tabular*}{\linewidth}{@{\extracolsep{\fill}}llllccccll@{}}
\toprule
\multicolumn{10}{@{}l}{\textbf{Structure stage} (400 views of 278 validation objects, true camera): support}\\ \midrule
Method & Data & Updates & Sampler & IoU$\uparrow$ & P & R & Thin R & IoU vs T & GPU-h \\ \midrule
12-step teacher & --- & --- & Euler & 0.597 & 0.725 & 0.727 & 0.476 & 1.000 & --- \\
25-step teacher & --- & --- & Euler & 0.597 & 0.724 & 0.727 & 0.472 & 0.973 & --- \\
12-step teacher, shift 5.0 & --- & --- & Euler & 0.598 & 0.725 & 0.727 & 0.480 & 0.978 & --- \\
Consistency & 5k (4,993) & 15,000 & re-noise & 0.592$\blacktriangledown$ & 0.724 & 0.718 & 0.460 & 0.796 & n/r (16) \\
Consistency & 10k (19,998) & 12,000 & re-noise & 0.596 & 0.728 & 0.722 & 0.464 & 0.797 & 27.8 (4) \\
Consistency & 10k (19,998) & 12,000 & Euler & 0.587$\blacktriangledown$ & 0.710 & 0.728 & 0.477 & 0.836 & 27.8 (4) \\
DMD2 (no GAN) & 10k (19,998) & 3,000 (15,000) & re-noise & 0.587$\blacktriangledown$ & 0.676 & 0.775 & 0.608 & 0.785 & 18.0 (4) \\
DMD2 (no GAN) & 10k (19,998) & 3,000 (15,000) & Euler & 0.586$\blacktriangledown$ & 0.675 & 0.773 & 0.620 & 0.825 & 18.0 (4) \\
MFD & 10k (19,998) & 1,000 (11,000) & re-noise & 0.599 & 0.736 & 0.718 & 0.453 & 0.812 & 19.0 (4) \\
MFD & 10k (19,998) & 1,000 (11,000) & Euler & 0.594$\blacktriangledown$ & 0.728 & 0.716 & 0.449 & 0.888 & 19.0 (4) \\
LFM & 10k (19,998) & 3,000 & Euler & 0.598 & 0.729 & 0.724 & 0.472 & 0.871 & 3.0 (4) \\
Trajectory & 5k (10,486) & 250 & Euler & 0.595$\blacktriangledown$ & 0.725 & 0.723 & 0.467 & 0.897 & 0.13 (1) \\
Trajectory (Ours) & 5k (10,486) & 500 & Euler & 0.598 & 0.727 & 0.726 & 0.471 & 0.894 & 0.25 (1) \\
Trajectory & matched (10,486) & 500 & Euler & 0.600 & 0.730 & 0.727 & 0.484 & 0.893 & 0.29 (4) \\
Trajectory & matched (10,486) & 3,000 & Euler & 0.597 & 0.727 & 0.726 & 0.470 & 0.874 & 1.8 (4) \\
Trajectory & 10k (19,998) & 500 & Euler & 0.599 & 0.729 & 0.727 & 0.469 & 0.891 & 0.29 (4) \\
Trajectory & 10k (19,998) & 3,000 & Euler & 0.597 & 0.725 & 0.727 & 0.468 & 0.868 & 1.8 (4) \\
Trajectory, seed 2 & 10k (19,998) & 500 & Euler & 0.598 & 0.728 & 0.727 & 0.486 & 0.890 & 0.29 (4) \\
Trajectory, seed 2 & 10k (19,998) & 3,000 & Euler & 0.599 & 0.728 & 0.728 & 0.484 & 0.859 & 1.8 (4) \\
\bottomrule
\end{tabular*}
\end{table}

Table~\ref{tab:distill_ablation_full} is the full version of Table~\ref{tab:distill}(a). It compares consistency distillation~\citep{rw_song2023consistency}, DMD2 without its GAN term~\citep{rw_yin2024dmd2}, mean-flow distillation (MFD)~\citep{rw_zhao2026mfd}, a Lagrangian flow-map objective (LFM)~\citep{rw_boffi2024flowmap} on the 4-step deployment grid, and trajectory distillation (Section~\ref{sec:method-distill}). MFD matches interval-averaged velocities, the quantity introduced by MeanFlow~\citep{ap_meanflow}, and about half of the training states of LFM come from the student's own rollouts, as in on-policy flow-map distillation~\citep{rw_gu2026anyflow}.
In Table~\ref{tab:distill_ablation_full} every student takes 4 steps, and the teacher uses guidance scale 7.5 and time shift 3.0. The column Sampler gives the sampler used at evaluation (re-noise: the sampler of consistency models), and the GPU-hours exclude the teacher-trajectory cache. There, the consistency and DMD2 students with the Euler sampler are also significantly lower than the 12-step teacher (Holm over the 18 runs). In Table~\ref{tab:distill}(b) the shape latents come from \sysname with the 4-step consistency structure student (5k set) in place of its structure student, and the texture-stage times are measured with the adapter as a separate branch.

\paragraph{Objective details.} In the consistency objective of Section~\ref{sec:method-distill}, the endpoint of the guided 12-step teacher is memoized during training. The endpoint $\tau=1$ of the discretization is itself not drawn, although the deployed 1-step students are evaluated there, and the Euler step of the frozen guided teacher is a live teacher call, not a point on the interpolation path. Equation~\eqref{eq:traj} relates to further prior objectives. TRACT~\citep{rw_berthelot2023tract} also distills many teacher steps into one student step, in one or two rounds. Equation~\eqref{eq:traj} also restricts the progressive objective of flow map matching~\citep{rw_boffi2024flowmap}, whose Lagrangian objective is one of those compared in Table~\ref{tab:distill_ablation_full}, and the consistency-trajectory objective~\citep{rw_kim2024ctm} to adjacent pairs of a fixed deployment grid. At deployment the texture student runs through the released sampler and the structure student through an equivalent conditional-only Euler loop on its non-uniform grid (checked against the released sampler). The grid of the structure student is a subset of the 12-step grid only at the time shift 3.0 of the shape and texture flows, not at the 5.0 of the released structure sampler.

\paragraph{Teacher trajectory caches.} Only trajectory distillation needs a cache: for each of its training views we run the 12-step teacher once (guided for the structure flow, unguided for the texture flow, both at time shift 3.0) and store the states at the segment boundaries. The structure cache stores the $16^3\times 8$ latent at eight of the 13 grid points per view, so that several candidate grids can be trained from one cache, and the texture cache stores per-view records conditioned on the teacher's own shape latent.

\paragraph{Training details and cost.}
A forward pass uses one view, and an update averages 16 views for the consistency students and 4 for the trajectory students. The moving average of the adapter has decay 0.99. The two consistency shape students and the 4-step consistency structure student trained on the 5k set used learning rate $2\times10^{-6}$ on 16 GPUs (updates: Section~\ref{sec:method-training} and Table~\ref{tab:distill_ablation_full}). The trajectory students use AdamW at learning rate $10^{-4}$ without weight decay and with gradient clipping at 10. The texture student was trained in two phases. Updates 1--1,500 ran on four GPUs, drawn from the 13,398 views that had teacher-trajectory records when the phase started. Updates 1,501--3,000 ran on one GPU with gradient accumulation over four views, drawn from all 19,998 views and resumed from the checkpoint after update 1,500 (adapter, moving average and optimizer state).
The rounding to bf16 when the adapters of the fine-shape and texture students are folded (Section~\ref{sec:method-model}) removes 48.2\% (fine shape) and 5.7\% (texture) of the Frobenius norm of the adapter update, and every number of the pipeline in this report is measured with the folded weights. Training the structure and texture students cost about 34 GPU-hours on H200 GPUs (per run: Table~\ref{tab:distill}), about 26.5 of them for the teacher-trajectory caches. Of the 19,998 texture records, 12,704 take the teacher's fine-shape latent from fine-shape teacher trajectories that a separate run stored in another software environment. Rebuilt on the current machine with unchanged code, every training cache is reproduced bit for bit on the views checked except these stored trajectories (relative difference growing to 6--9\% at the last step on the two views checked), so the 12,704 records cannot be rebuilt exactly.

\paragraph{Shape students retrained on the 10k set.}
The coarse- and fine-shape students were also retrained on the 10k set, with the checkpoints (3,000 and 2,000 updates) chosen on the 500-object Toys4K development part by a rule fixed in advance. On the 400 validation objects the retrained pair is slightly worse than the deployed students on every normal-map metric (normal PSNR $-0.08$\,dB $[-0.10, -0.07]$, mean angular error $+0.33^\circ$ $[+0.29, +0.37]$), slightly lower in Chamfer distance ($-0.06\times10^{-3}$ $[-0.11, -0.02]$) and unchanged in F-score, silhouette IoU and E2E time. We therefore keep the deployed students.

\section{Techniques Tried and Not Adopted}
\label{app:ledger}

The techniques below were tried and not adopted, whereas the adopted ones are the rows of Table~\ref{tab:optimizations}. Results are on the validation data unless an entry names another set.

\paragraph{Attention that is neither sparse nor low-rank.}
Block-sparse attention with fixed spatial blocks or learned content-based blocks (texture and fine-shape flows) is not faster at accurate settings: the attention is not local, and selecting the blocks costs what it saves. Linear attention with per-layer feature maps distilled from the released attention (texture flow) has a relative error per layer of 0.10--0.60, or 0.02--0.35 with three exact heads kept, on objects of the 5k set not used for training. Hence, the attention is not low-rank either. 8-bit floating-point attention gives a large latent error and is not faster. A sparse-attention texture student, with the attention mask learned by re-distillation, was rejected: it was far from the dense student after 600 updates, and the masked attention saves time only on heavy objects.

\paragraph{Little skippable depth, few mergeable tokens.}
In the texture flow few layers can be skipped or tokens merged, and the feed-forward layers are small and densely active (layer skipping, token merging, sparse feed-forward activations). Distilling the VAE decoders to half their blocks was rejected: the decoders are bound by kernel launches, so halving the blocks gives too small a speed-up.

\paragraph{Few removable steps without training.}
The 2-step texture student is used instead of fewer Euler steps (8, 6, 4), Adams--Bashforth multistep solvers and step caching for the texture flow (Section~\ref{sec:method-steps}).

\paragraph{A thresholded support that any perturbation changes.}
FlashAttention-3 and adapter folding in the structure and coarse-shape stages change the sample (Sections~\ref{sec:method-infra-flows} and~\ref{sec:method-model}), so neither is used there (Table~\ref{tab:optimizations}). Approximate sharing of the conditioning across stages (pooled features or backbone tokens) does not shorten the E2E time, and its stronger variants change the coarse-shape support.

\paragraph{GPU contention that offsets further overlap.}
Computing the conditioning of later stages ahead on a side stream leaves the output unchanged but does not shorten the E2E time, because the work moves into the preceding stage. Other schedules of the mesh pre-processing under the texture flow leave the output unchanged and are not faster either, since the GPU work of the two is fixed and they share one GPU. Overlapping shape decoding with the mesh pre-processing is not possible, because the pre-processing needs the decoded mesh. A per-object cache of pre-processing, structure conditioning and structure latent helps only when the same object is requested again, and a cache of the texture conditioning projections alone is not measurably faster.

\paragraph{Launch-bound or non-deterministic decoders and mesh code.}
Compiling or fusing the non-attention operations of the fine-shape and texture flows costs minutes of compilation per process, and the approximate arithmetic worsened the shape metrics. Caching sequence metadata and timestep embeddings to avoid host synchronization leaves the output unchanged with a tiny gain and was not kept. A static CUDA graph fails to capture the coarse-shape student, whose sparse-convolution library tunes its kernels at run time, and gains little for the structure decoder. The VAE decoders are bound by kernel launches: exact fusion or graph capture gains little, and compiled fusion is approximate and slow to compile. In the GPU simplification, deterministic modes are slower and graph capture conflicts with the concurrent texture thread.

\paragraph{Mesh post-processing alternatives.}
The other schedules of the CPU unwrap (a single atlas in parallel, $2\times2$ tiles, two or four inner threads) were not faster than one thread per tile on 16 tiles. Other GPU atlas variants (split angles, merging of small charts, keeping the released chart computation) gain nothing over the adopted single-pass chart segmentation, and keeping the released chart computation keeps its cost. Coarser remeshing (fixed 1/2, 1/3 and 1/4 grids, a face budget, octree-adaptive grids, a coarse iso-surface) loses thin detail (halving the distance band of the adopted remesh: Appendix~\ref{app:perception_gap}). Among export variants, uncompressed PNG multiplies the file size and lossless WebP encodes far more slowly.

\paragraph{Other students and quality attempts.}
The alternative students of Tables~\ref{tab:distill_ablation_full}, \ref{tab:distill} and~\ref{tab:ablation} did not improve on the deployed ones (Section~\ref{sec:ablations}). A further texture student, trained for 4 steps and sampled with 2, has an L1 distance to the 12-step teacher of $3.26\times10^{-2}$ against $3.25\times10^{-2}$ for the 2-step student trained on the same set.
Mean-flow distillation of the texture student with an auxiliary network raised the DINOv2 Fr\'echet distance to the ground-truth renders by 73.7 after 1,000 updates, relative to the 2-step texture student (39 objects of the 40-object subset): it overshoots the fine detail, restoring the high-frequency content beyond the ground-truth level. The following were judged during development and not measured on the validation data. Consistency distillation of the texture flow (4 steps) has a larger distance to the 12-step output than four Euler steps of the teacher. One joint 1-step student for fine shape and texture is far less accurate than the separate students in both outputs. Mean-flow distillation for the coarse- and fine-shape students was stopped, with no ground-truth gain at a high training cost (about 660 GPU-h for the fine-shape run). DMD2 with a GAN term for a 6-step coarse-shape student was stopped because the discriminator stayed at chance, no different from consistency distillation. Training the coarse-shape student on supports of the structure student instead of the teacher's had no detectable effect, and fine-tuning the structure flow on ground-truth occupancy (cross-entropy and Dice losses through the frozen decoder) was rejected because the loss falls while the support inflates. The following were not evaluated: truncated consistency models~\citep{rw_lee2025tcm} for the structure student, a uniform 4-step grid for the structure trajectory student, and 1-step and 4-step texture students sampled at their own step count. Also not evaluated were a 4-step consistency student for the fine-shape stage, a 2-step consistency student for the coarse-shape stage, a cap on the fine-shape resolution, remeshing disabled, and further pilots for improving the structure stage.

\end{document}